%% file: main.tex
\documentclass[10pt,twocolumn,letterpaper]{article}

\usepackage{iccv}
\usepackage{times}
\usepackage{epsfig}
\usepackage{graphicx}
\usepackage{amsmath}
\usepackage{amssymb}

\usepackage{enumitem} 
\usepackage{array,multirow,makecell,tabularx}
\usepackage{color}
\usepackage{comment}
\usepackage{xcolor}
\usepackage{xspace}
\usepackage{footnote}
\usepackage{caption}

\usepackage{array} 

\usepackage{booktabs}
\usepackage{float}
\usepackage{stfloats}

\usepackage[pagebackref=true,breaklinks=true,letterpaper=true,colorlinks,bookmarks=false]{hyperref}

\iccvfinalcopy 

\def\iccvPaperID{1744} 

\ificcvfinal\fi

\newcommand{\arch}[1]{\textsc{#1}}

\newcommand{\Fref}[1]{Figure~\ref{#1}}
\newcommand{\Sref}[1]{Section~\ref{#1}}
\newcommand{\Tref}[1]{Table~\ref{#1}}

\newcommand{\Eclass}{E_{C}\xspace}  %
\newcommand{\Estyle}{E_{S}\xspace}  %

\newcommand{\x}{\mathbf{x}\xspace}  
\newcommand{\xt}{\tilde{\x}\xspace}  
\newcommand{\xp}{\mathbf{x^+}\xspace}  

\newcommand{\s}{\mathbf{s}\xspace}  
\newcommand{\st}{\tilde{\s}\xspace}  
\newcommand{\spp}{\mathbf{s^+}\xspace}  
\newcommand{\sn}{\mathbf{s}_i^-\xspace}  

\newcommand{\stt}{\s'\xspace}

\newcommand{\p}{\mathbf{p}\xspace}  
\newcommand{\pp}{\mathbf{p^+}\xspace}  
\newcommand{\Pij}{\mathbf{P}_{ij}\xspace}
\renewcommand{\P}{\mathbf{P}\xspace}  

\newcommand{\y}{y\xspace} 
\newcommand{\yt}{\tilde{\y}\xspace} 

\newcommand{\rchi}{\raisebox{2pt}{$\chi$}} 

\newcommand{\tablestyle}[2]{\setlength{\tabcolsep}{#1}\renewcommand{\arraystretch}{#2}\centering\footnotesize}
\newlength\savewidth\newcommand\shline{\noalign{\global\savewidth\arrayrulewidth
  \global\arrayrulewidth 1pt}\hline\noalign{\global\arrayrulewidth\savewidth}}

\include{figures}

\include{table}

\begin{document}

\title{Appendix: Rethinking the Truly Unsupervised Image-to-Image Translation}

\author{Kyungjune Baek\thanks{Work done during his internship at Clova AI Research.}\\
Yonsei University\\
{\tt\small bkjbkj12@yonsei.ac.kr}
\and
Yunjey Choi\\
NAVER AI Lab\\
{\tt\small yunjey.choi@navercorp.com}
\and
Youngjung Uh\\
Yonsei University\\
{\tt\small yj.uh@yonsei.ac.kr}
\and
Jaejun Yoo\\
UNIST\\
{\tt\small jaejun.yoo@unist.ac.kr}
\and
Hyunjung Shim\thanks{Hyunjung Shim is a corresponding author.}\\
Yonsei University\\
{\tt\small kateshim@yonsei.ac.kr}
}

\maketitle
\ificcvfinal\thispagestyle{empty}\fi

\input{0.abstract}
\input{1.introduction}
\input{2.related_work}
\input{3.method}
\input{4.experiments}

\input{5.conclusion}
{\small \noindent\textbf{Acknowledgements.}
All experiments were conducted on NAVER Smart ML (NSML)~\cite{nsml} platform. This research was supported by the NRF Korea funded by the MSIT (NRF-2019R1A2C2006123), the IITP grant funded by the MSIT (2020-0-01361, YONSEI University, 2020-0-01336, Artificial Intelligence graduate school support (UNIST)), and the Korea Medical Device Development Fund grant (Project Number:  202011D06).}

{\small
\bibliographystyle{ieee_fullname}

}

\newpage
\input{6.appendix}

\end{document}

%% file: figures.tex
\newcommand{\figIntro}{
\begin{figure}[t!]
\centering
\includegraphics[width=\linewidth]{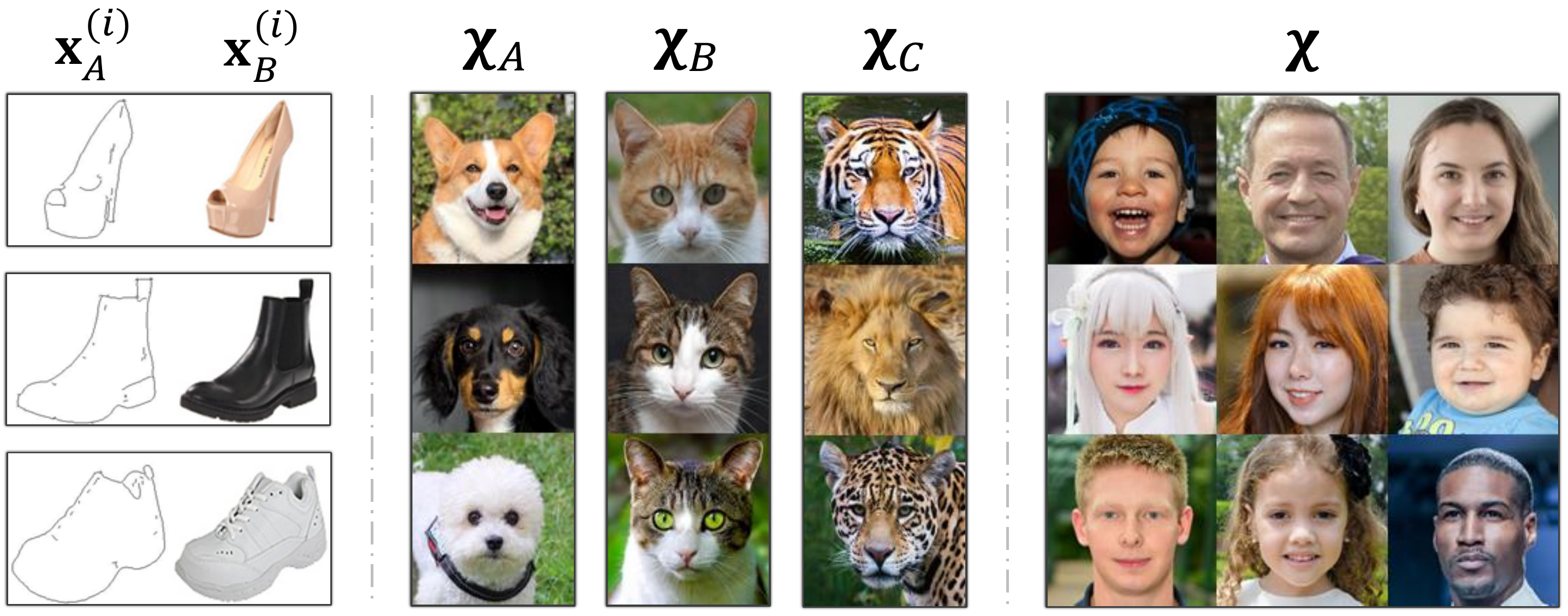}
\vspace{-0.1cm}
\makebox[0.2\linewidth][c]{\footnotesize{(a) Image-level}}\hfill
\makebox[0.45\linewidth][c]{\footnotesize{(b) Set-level}}\hfill
\makebox[0.35\linewidth][c]{\footnotesize{(c) Unsupervised}}\hfill \\
\caption{\small{\textbf{Levels of supervision.} To perform image-to-image translation, existing methods need either \textbf{(a)} a dataset with input-output pairs or, \textbf{(b)} a dataset with domain information. Our method is capable of learning mappings among multiple domains using \textbf{(c)} a dataset without any supervision.}} 

\label{fig:introduction}
\vspace{-0.3cm}
\end{figure}
}

\newcommand{\figModel}{
\newcommand{\h}{0.9\columnwidth}
\begin{figure}[t!]
\centering
\includegraphics[width=\h]{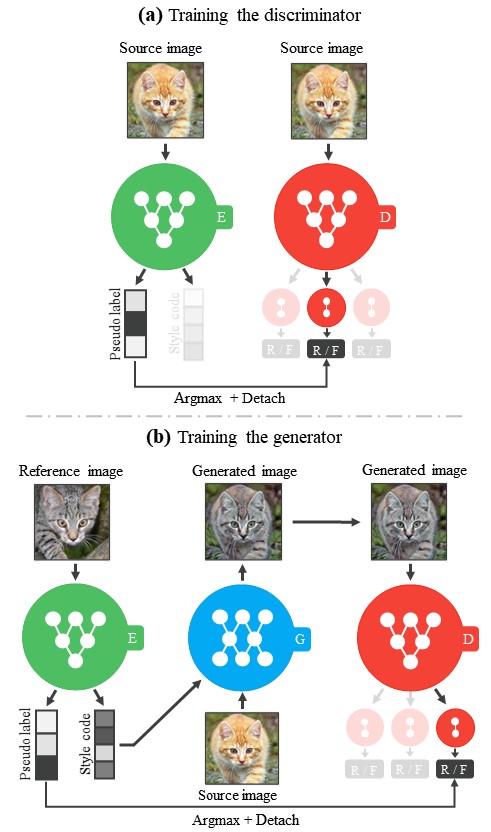}\\
\caption{\small{\textbf{Overview of our proposed method.} The figure illustrates how our model changes the breed of a cat. \textbf{(a)} Our guiding network $E$ estimates the domain and use it to train the multi-task discriminator $D$. \textbf{(b)} Both the style code and the estimated domain of a reference image is used for training the generator $G$.}
}
\label{fig:model}
\vspace{-0.5 cm}
\end{figure}
}

\newcommand{\figmaincomparison}{
\renewcommand{\h}{14.0mm}
\newcommand{\hout}{13.56mm}
\begin{figure*}[t]
\centering
\footnotesize{
\makebox[\h][c]{Source}\hspace{-0.3mm}
\makebox[\h][c]{Reference}\hspace{0.9mm}
\makebox[\hout][c]{\arch{\normalsize a}}\hspace{0.0mm}
\makebox[\hout][c]{\arch{\normalsize b}}\hspace{0.0mm}
\makebox[\hout][c]{\arch{\normalsize c}}\vspace{0.0mm}
\makebox[\hout][c]{\arch{\normalsize d}}\hspace{0.0mm}
\makebox[\hout][c]{\arch{\normalsize e}}\hspace{0.0mm}
\makebox[\hout][c]{\arch{\normalsize f}}\vspace{0.0mm}
\makebox[\hout][c]{\arch{\normalsize g}}\hspace{0.0mm}
\vspace{0.3mm}
}\\\
\includegraphics[width=130mm]{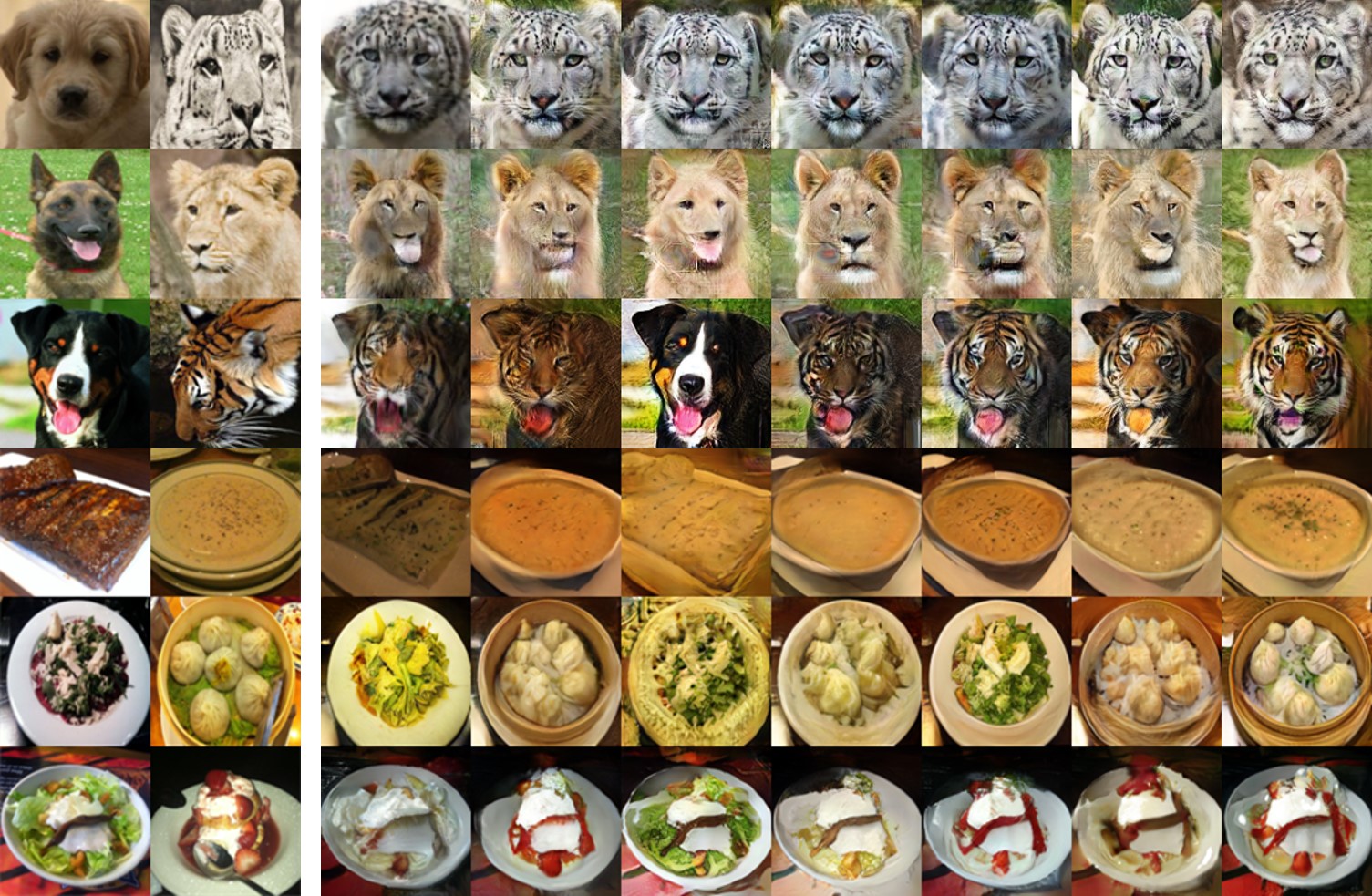}
\vspace{-2mm}
\caption{\small{Qualitative comparison of translation results using each configuration in \Tref{tab:main}. 
Here, \texttt{B} reflects the style feature (\eg species or type of food) of the reference images while \texttt{A} does not. 
The model \texttt{C} performs much worse than \texttt{A} and \texttt{B} in that it overly adopts the source image, not adequately merging styles and contents from both sides. The model \texttt{D} generates more plausible images than \texttt{C} but fails to reflect the characteristics of the reference images. For example, \texttt{D} on fifth row does not look like \emph{several pieces of dumpling} due to its shape and dish color, meaning that the reference styles are not properly reflected. 
Similarly, \texttt{E} also fails to generate the dumpling in the fifth row. 
TUNIT with sequential training \texttt{F} reflects the visual features of each reference on both datasets. However, in terms of visual fidelity, we observe that \texttt{G} consistently  outperforms \texttt{F}. Akin to the quantitative results, TUNIT achieves equivalent or even better visual quality than the set-level supervised model \texttt{A} and \texttt{B}.}
}
\vspace{-4mm}
\label{fig:main_comparison}
\end{figure*}
}

\newcommand{\figKcomparison}{
\renewcommand{\h}{14.0mm}
\renewcommand{\hout}{13.56mm}
\begin{figure*}[t!]
\centering
\vspace{-2mm}
\footnotesize{
\makebox[\h][c]{Source}\hspace{0.0mm}
\makebox[\h][c]{Reference}\hspace{0.7mm}
\makebox[\hout][c]{$\hat{K}$=1}\hspace{0.0mm}
\makebox[\hout][c]{$\hat{K}$=4}\hspace{0.0mm}
\makebox[\hout][c]{$\hat{K}$=7}\hspace{0.0mm}
\makebox[\hout][c]{$\hat{K}$=10}\hspace{0.0mm}
\makebox[\hout][c]{$\hat{K}$=13}\hspace{0.0mm}
\makebox[\hout][c]{$\hat{K}$=16}\hspace{0.0mm}
\makebox[\hout][c]{$\hat{K}$=20}\hspace{0.0mm}
}\\\
\includegraphics[width=130mm]{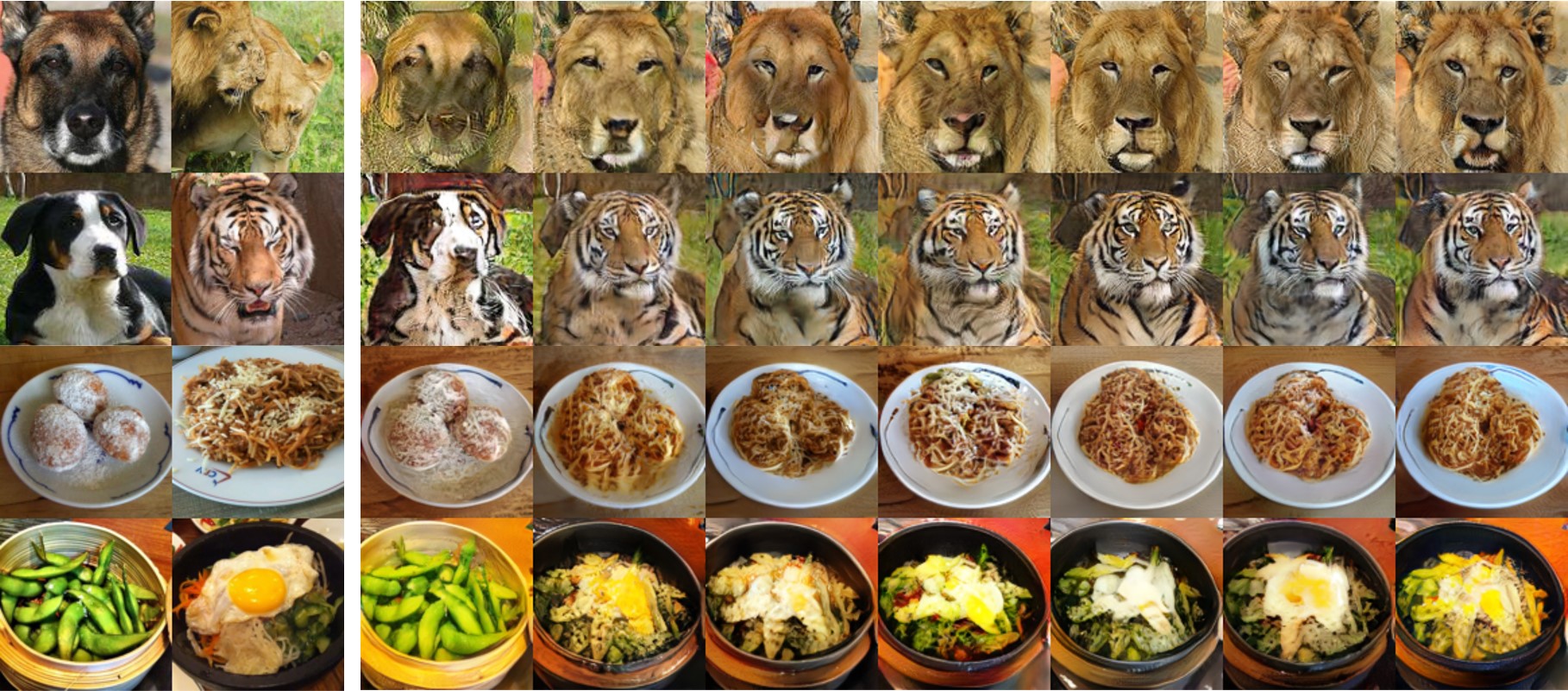}
\vspace{-2mm}
\caption{\small{Qualitative comparison on the number of pseudo domains $\hat{K}$. The performance varies along with $\hat{K}$. When we set $\hat{K}$ large enough, the results are reasonable.}}
\label{fig:k_comparison}
\end{figure*}
}

\newcommand{\figFFHQfew}{
\begin{figure}[t]
\centering
\includegraphics[width=0.6\columnwidth]{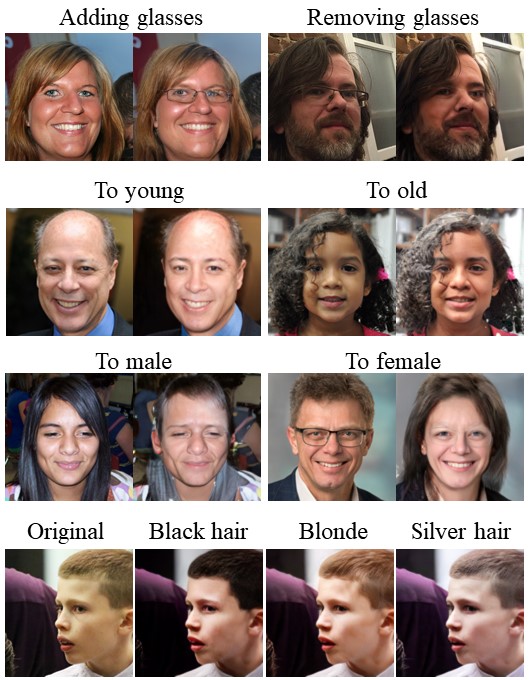}
\caption{\small{\textbf{Cross-domain attribute translation using 0.1\% of labeled samples.} As a practical application, we also apply TUNIT to FFHQ with few labels as a form of cross-domain translation. For each attribute, we train TUNIT separately (there are four models.). To this end, we manually label 35 images for each domain -- one contains the attribute and another does not contain it. Then, we train TUNIT with 70 labeled samples (0.1\% of the dataset) and remaining unlabeled samples. To add or remove the attribute, we use the average style vector of each domain. This result shows that TUNIT can greatly reduce the labeling cost.}}
\label{fig:ffhqfew}
\vspace{-0.6cm}
\end{figure}
}

\newcommand{\figUnlabeled}{
\renewcommand{\h}{13.3mm}
\newcommand{\hsrc}{1.33mm}
\newcommand{\hdataset}{1.78mm}
\begin{figure*}[t]
\centering
\hspace{1mm}
\makebox[\h][c]{Source}\hspace{4mm}
\makebox[\h][c]{Reference}\hspace{5mm}
\makebox[\h][c]{FUNIT}\hspace{3mm}
\makebox[\h][c]{TUNIT}\hspace{10mm}
\makebox[\h][c]{Source}\hspace{3mm}
\makebox[\h][c]{Reference}\hspace{5mm}
\makebox[\h][c]{FUNIT}\hspace{3mm}
\makebox[\h][c]{TUNIT}\hspace{0mm}
\vspace{0.5mm}
\\
\rotatebox{90}{\makebox[52mm][c]{\textsc{afhq cat}}}\vspace{1mm}\hspace{0.5mm}
\includegraphics[width=0.2\textwidth]{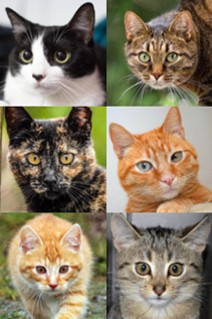}\hspace{\hsrc}
\includegraphics[width=0.2\textwidth]{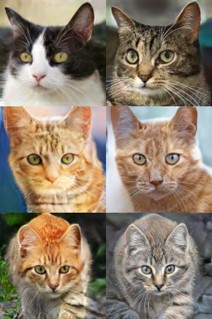}\hspace{\hdataset}
\rotatebox{90}{\makebox[52mm][c]{\textsc{ffhq}}}\vspace{0mm}
\includegraphics[width=0.2\textwidth]{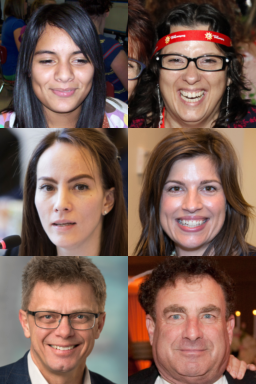}\hspace{\hsrc}
\includegraphics[width=0.2\textwidth]{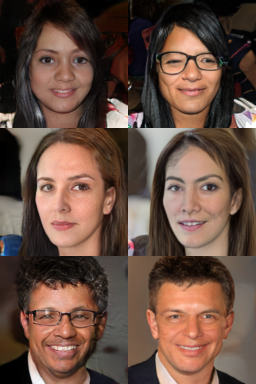}
\vspace{-1.2mm}
\\
\rotatebox{90}{\makebox[52mm][c]{\textsc{afhq wild}}}\vspace{1mm}\hspace{0.5mm}
\includegraphics[width=0.2\textwidth]{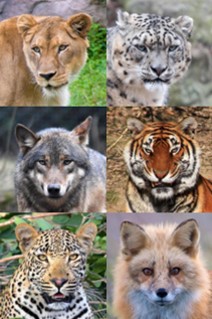}\hspace{\hsrc}
\includegraphics[width=0.2\textwidth]{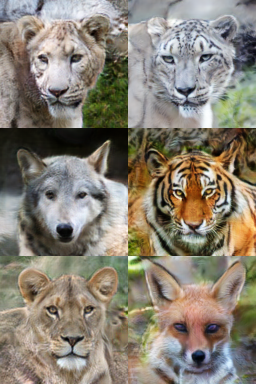}\hspace{\hdataset}
\rotatebox{90}{\makebox[52mm][c]{\textsc{lsun car}}}\vspace{0mm}\hspace{0.5mm}
\includegraphics[width=0.2\textwidth]{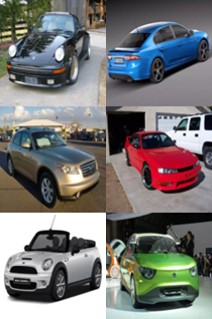}\hspace{\hsrc}
\includegraphics[width=0.2\textwidth]{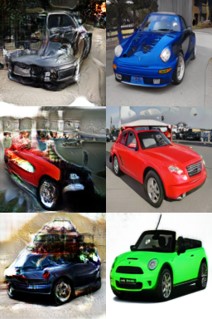}
\vspace{-0.4cm}
\caption{\small{Reference-guided image translation results on unlabeled datasets.}}
\label{fig:unlabeled_reference_guided}
\vspace{-0.5cm}
\end{figure*}
}

\newcommand{\figtsnecompare}{
\begin{figure}[t!]
\centering
\includegraphics[width=1.0\columnwidth]{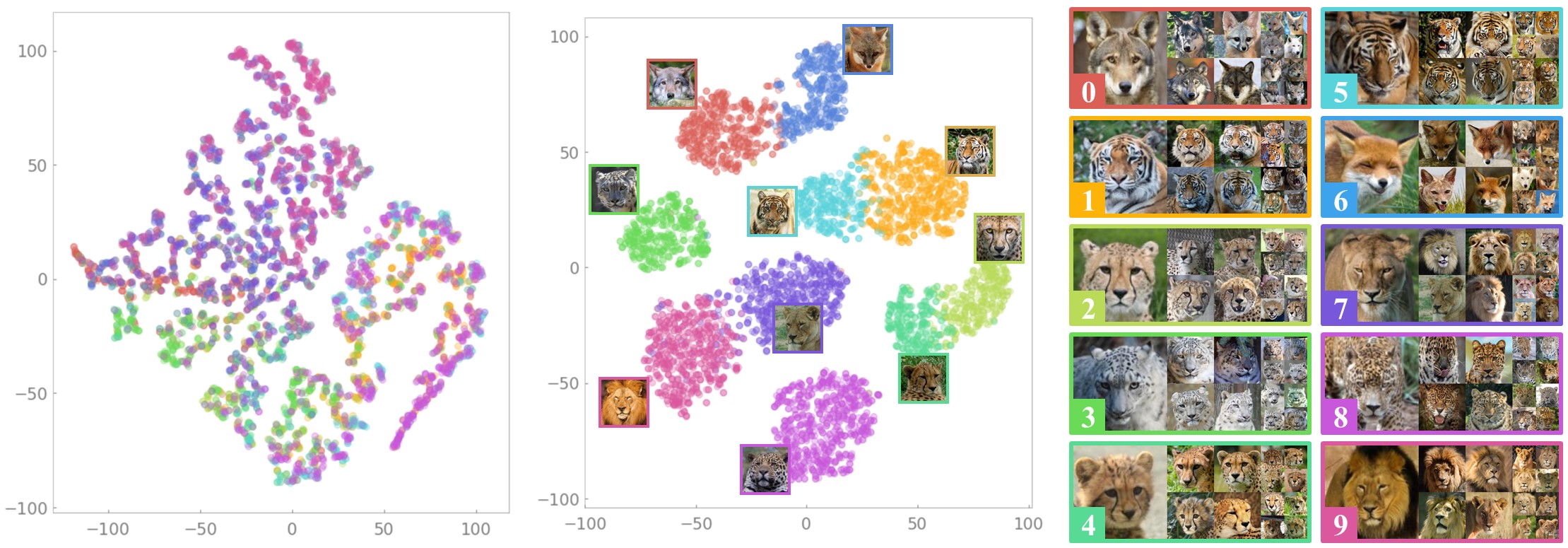}

\makebox[0.68\columnwidth][c]{\small{(a) t-SNE of FUNIT (L) / TUNIT (R)}}
\makebox[0.31\columnwidth][l]{\small{(b) Example images}}\hfill
\caption{t-SNE visualization of style space trained on AFHQ Wild. Since AFHQ Wild does not have ground-truth labels, each point is colored with the guiding network's prediction. Although we set the number of domains to be larger ($\hat{K}=10$) than it seems to have, the network practically creates six clusters by closely locating clusters of one species. 
}
\label{fig:tsne_compare}
\vspace{-0.4cm}
\end{figure}
}

\newcommand{\figTsneAfhqCat}{
\begin{figure}[!h]
\centering
\vspace{-10mm}
\includegraphics[width=0.96\linewidth]{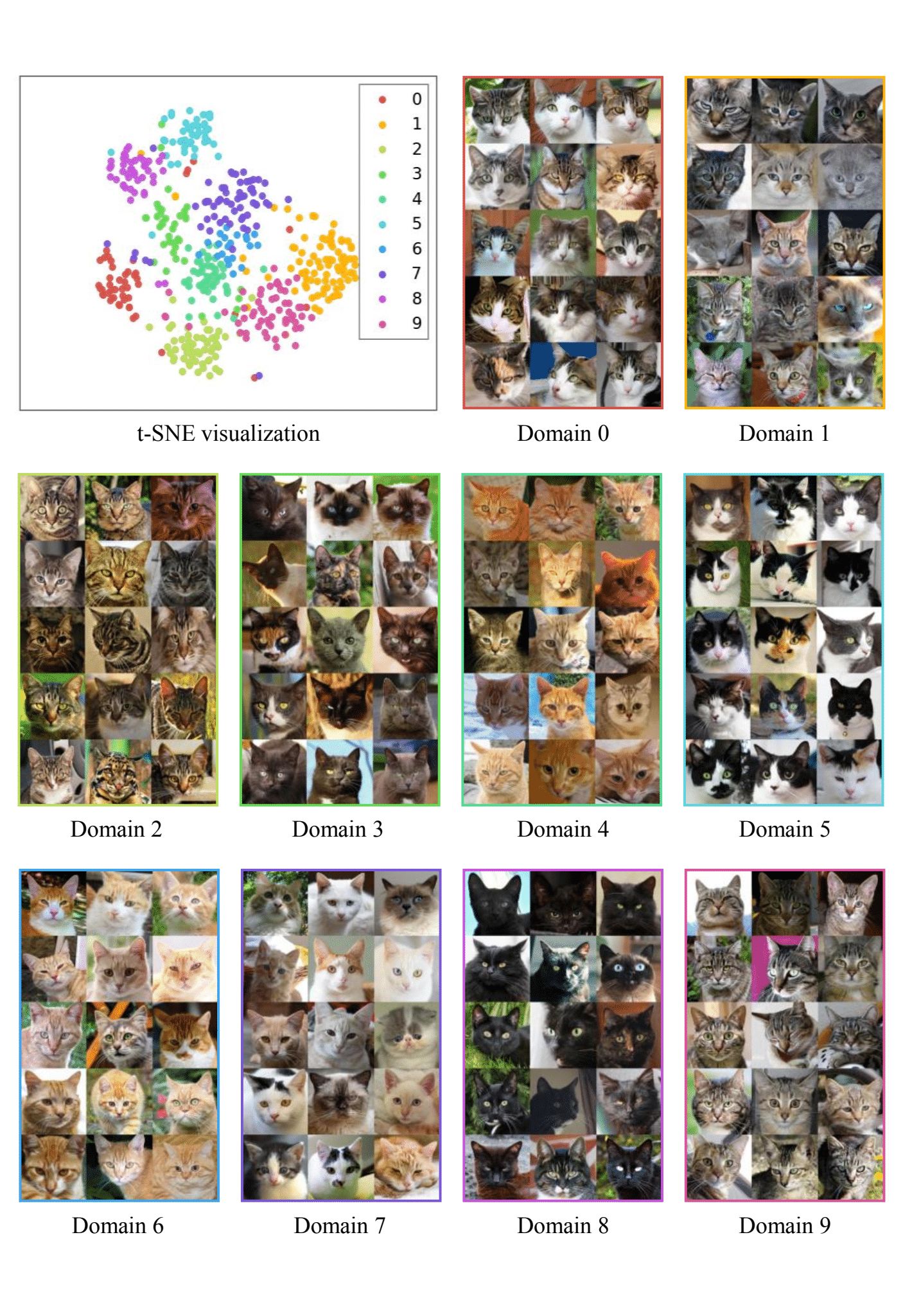}
\vspace{-10mm}
\caption{\textbf{t-SNE visualization and representative images of each domain for AFHQ Cat.}}
\label{fig:tsne_afhq_cat}
\end{figure}
}

\newcommand{\figTsneAfhqDog}{
\begin{figure}[!h]
\centering
\vspace{1.13cm}
\includegraphics[width=0.96\linewidth]{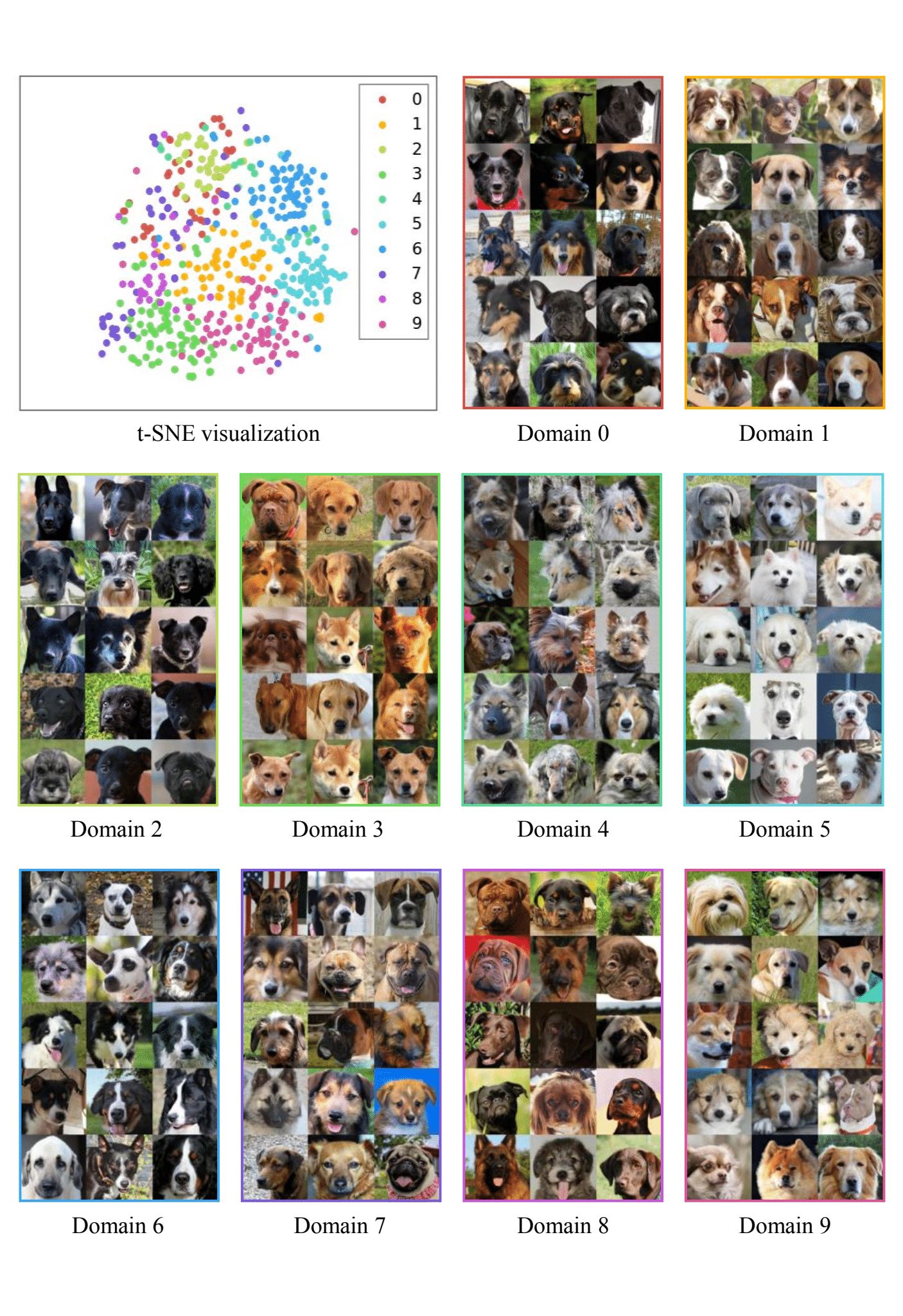}
\vspace{-10mm}
\caption{\textbf{t-SNE visualization and representative images of each domain for AFHQ Dog.}}
\label{fig:tsne_afhq_dog}
\end{figure}
}

\newcommand{\figTsneFfhq}{
\begin{figure}[!h]
\centering
\vspace{-10mm}
\includegraphics[width=0.96\linewidth]{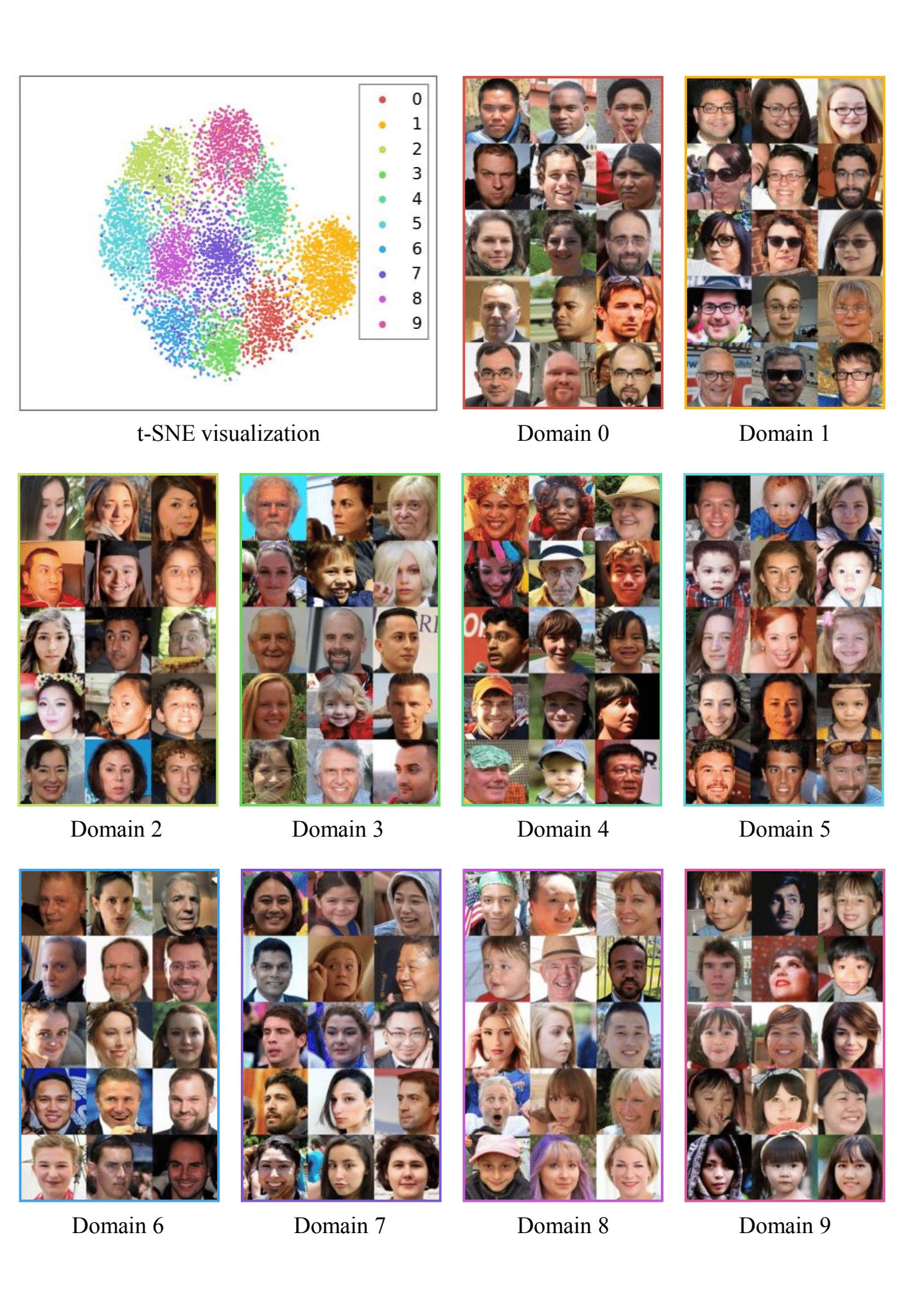}
\vspace{-10mm}
\caption{\textbf{t-SNE visualization and representative images of each domain for FFHQ.}}
\label{fig:tsne_ffhq}
\end{figure}
}

\newcommand{\figTsneLsun}{
\begin{figure}[!h]
\centering
\vspace{-10mm}
\includegraphics[width=0.96\linewidth]{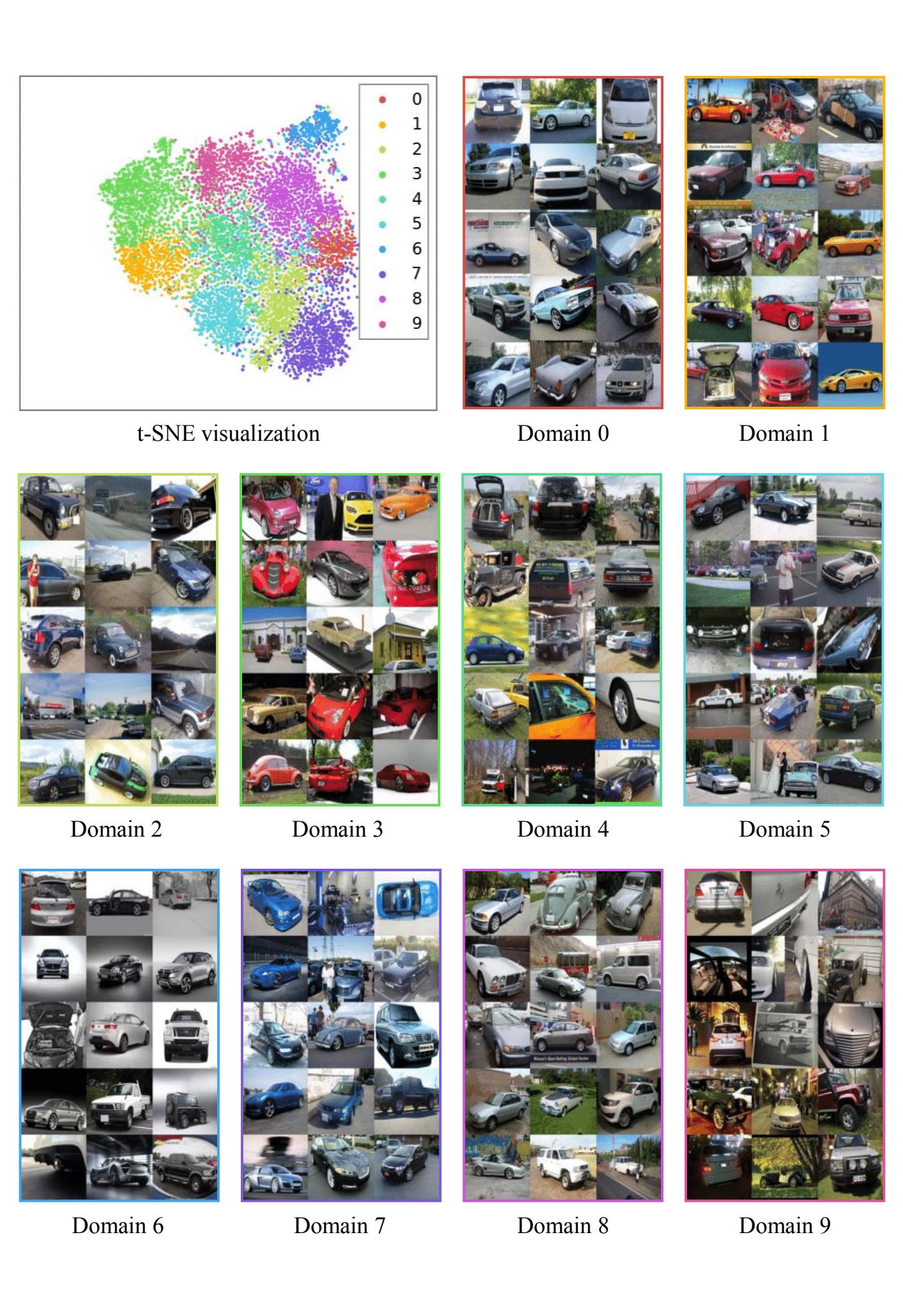}
\vspace{-10mm}
\caption{\textbf{t-SNE visualization and representative images of each domain for LSUN Car.}}
\label{fig:tsne_lsun}
\end{figure}
}

\newcommand{\figAppAnimalFaces}{
\begin{figure*}[b]
\centering
\vspace{-7mm}
\includegraphics[width=0.75\linewidth]{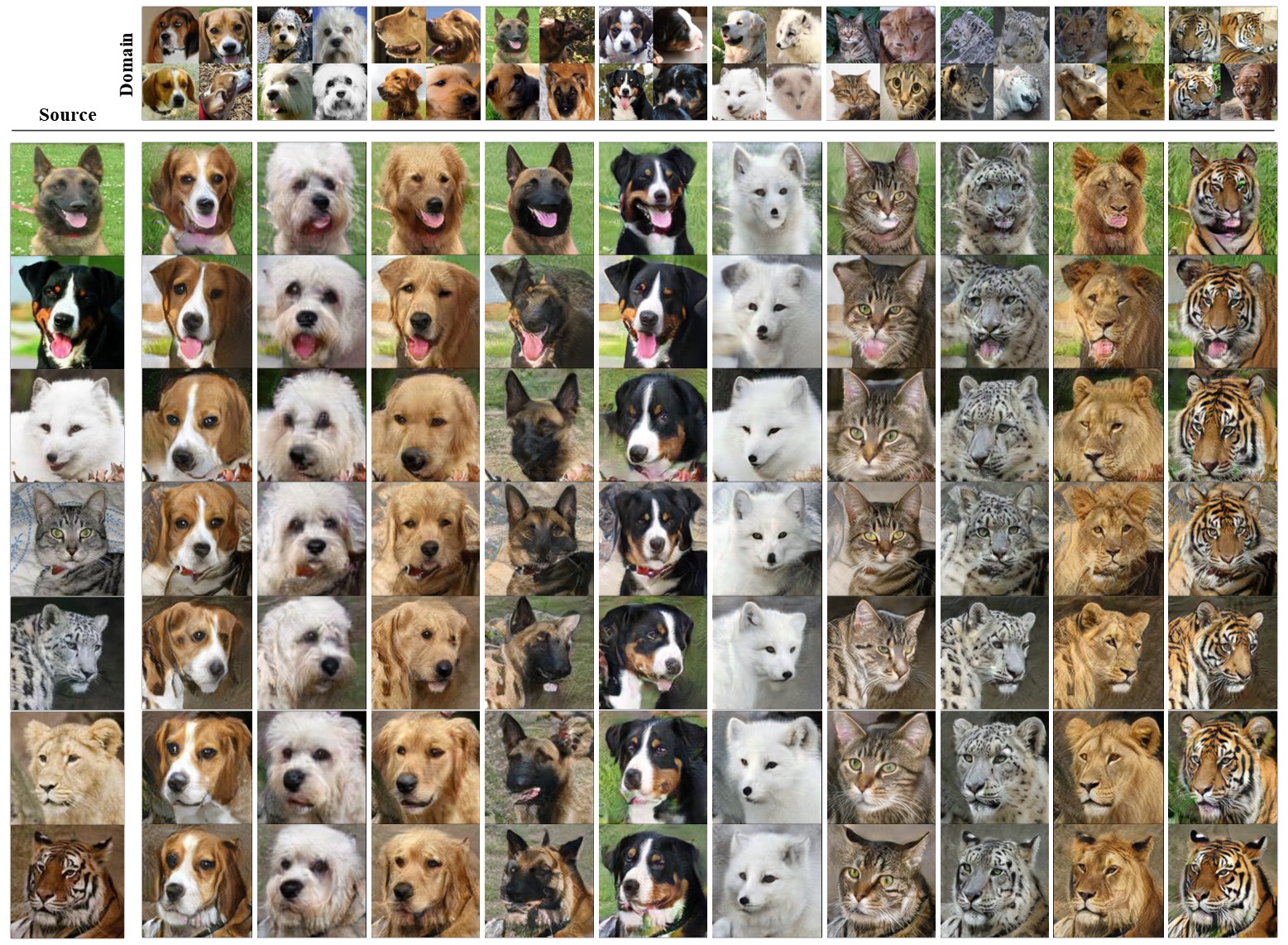}\\
\makebox[0.8\linewidth][c]{\textbf{(a) Results guided by average style vectors}}\vspace{1mm}
\includegraphics[width=0.75\linewidth]{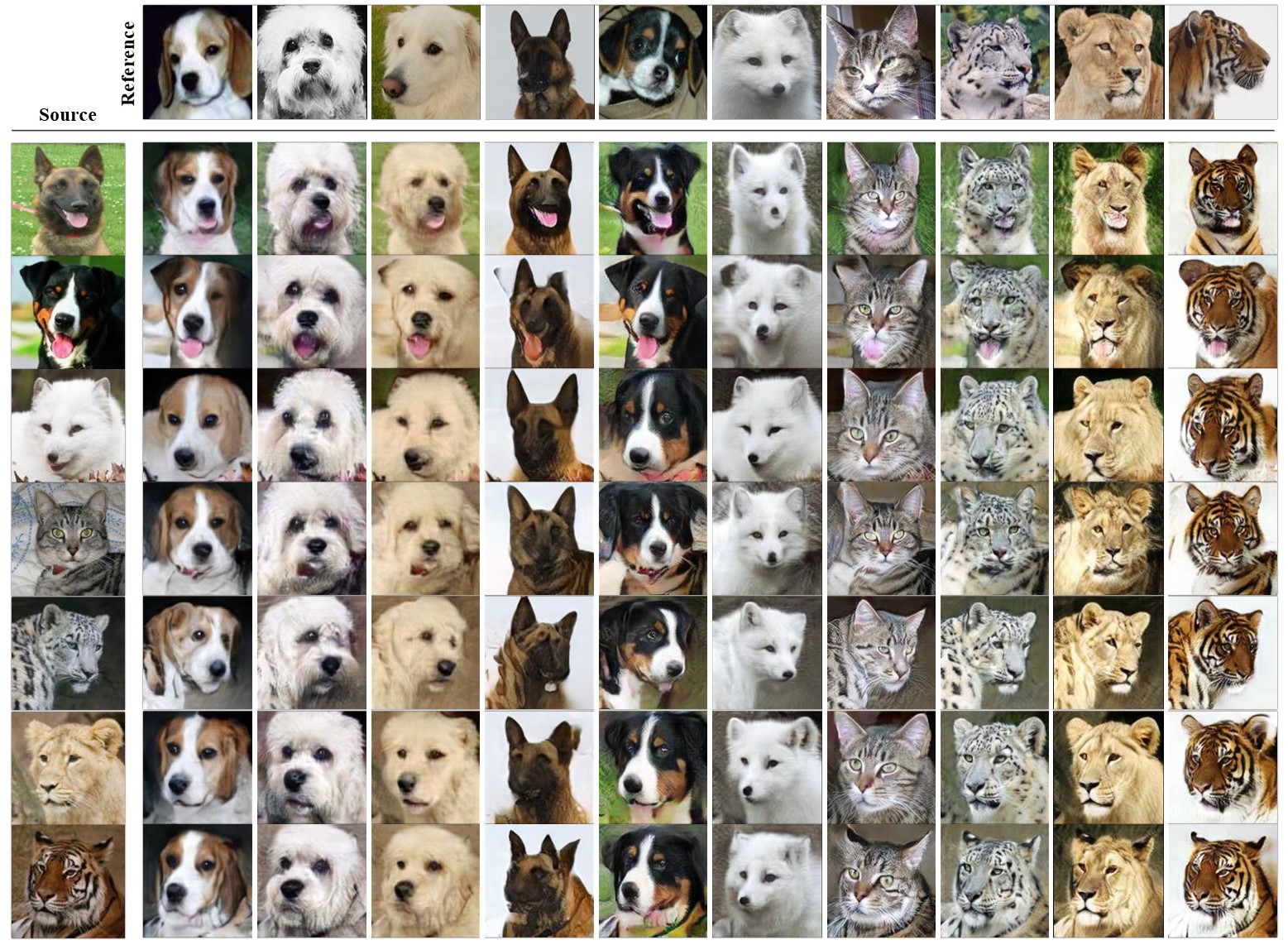}\\
\makebox[0.8\linewidth][c]{\textbf{(b) Results guided by reference images}}
\vspace{-3mm}
\caption{AnimalFaces-10, unsupervised image-to-image translation results.}
\label{fig:AppAnimalFaces}
\end{figure*}
}

\newcommand{\figAppAFHQCat}{
\begin{figure*}[b]
\centering
\vspace{-7mm}
\includegraphics[width=0.75\textwidth]{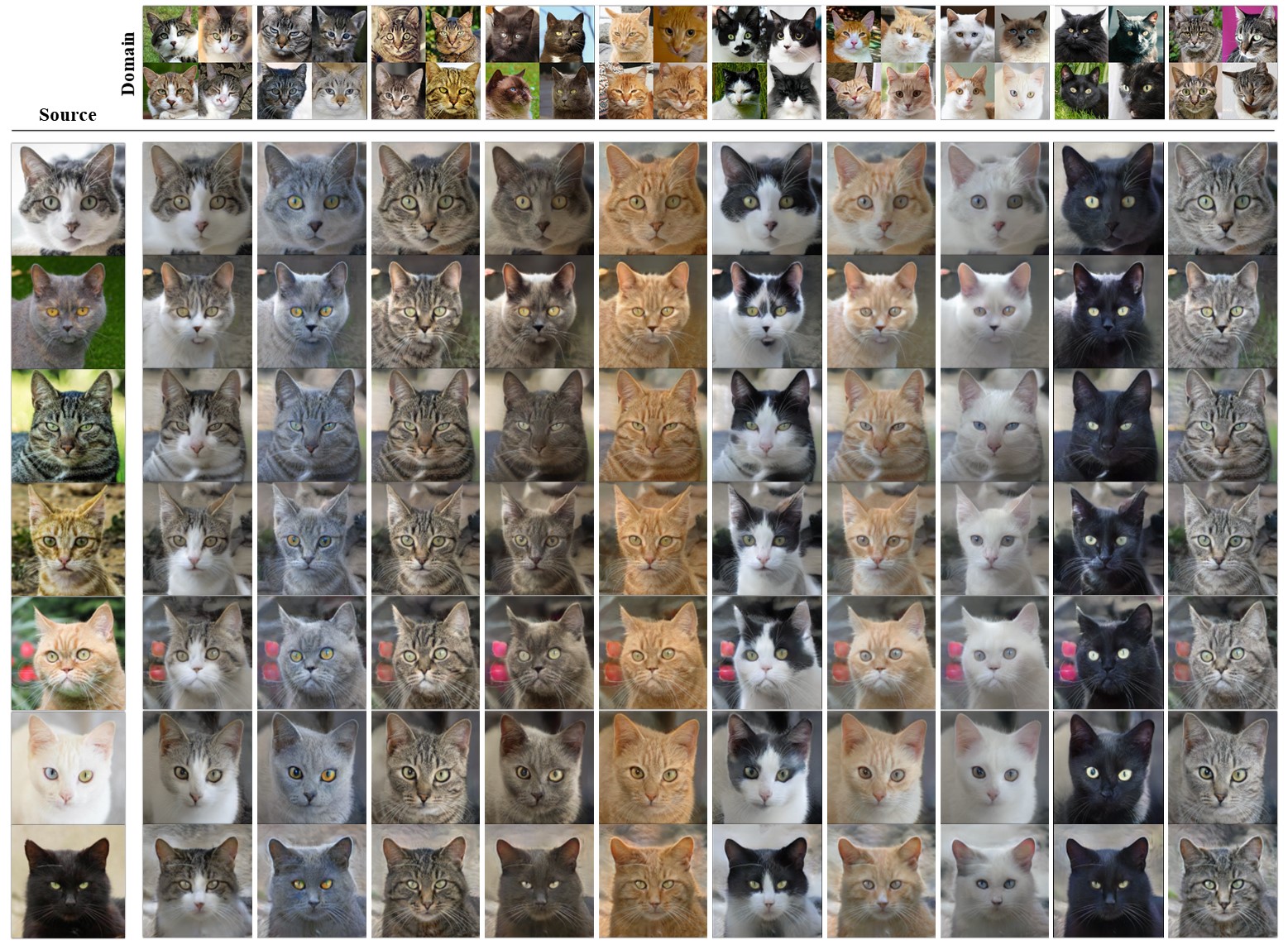}\\
\makebox[1\linewidth][c]{\textbf{(a) Results guided by the average style code of each domain}}\vspace{1mm}
\includegraphics[width=0.75\textwidth]{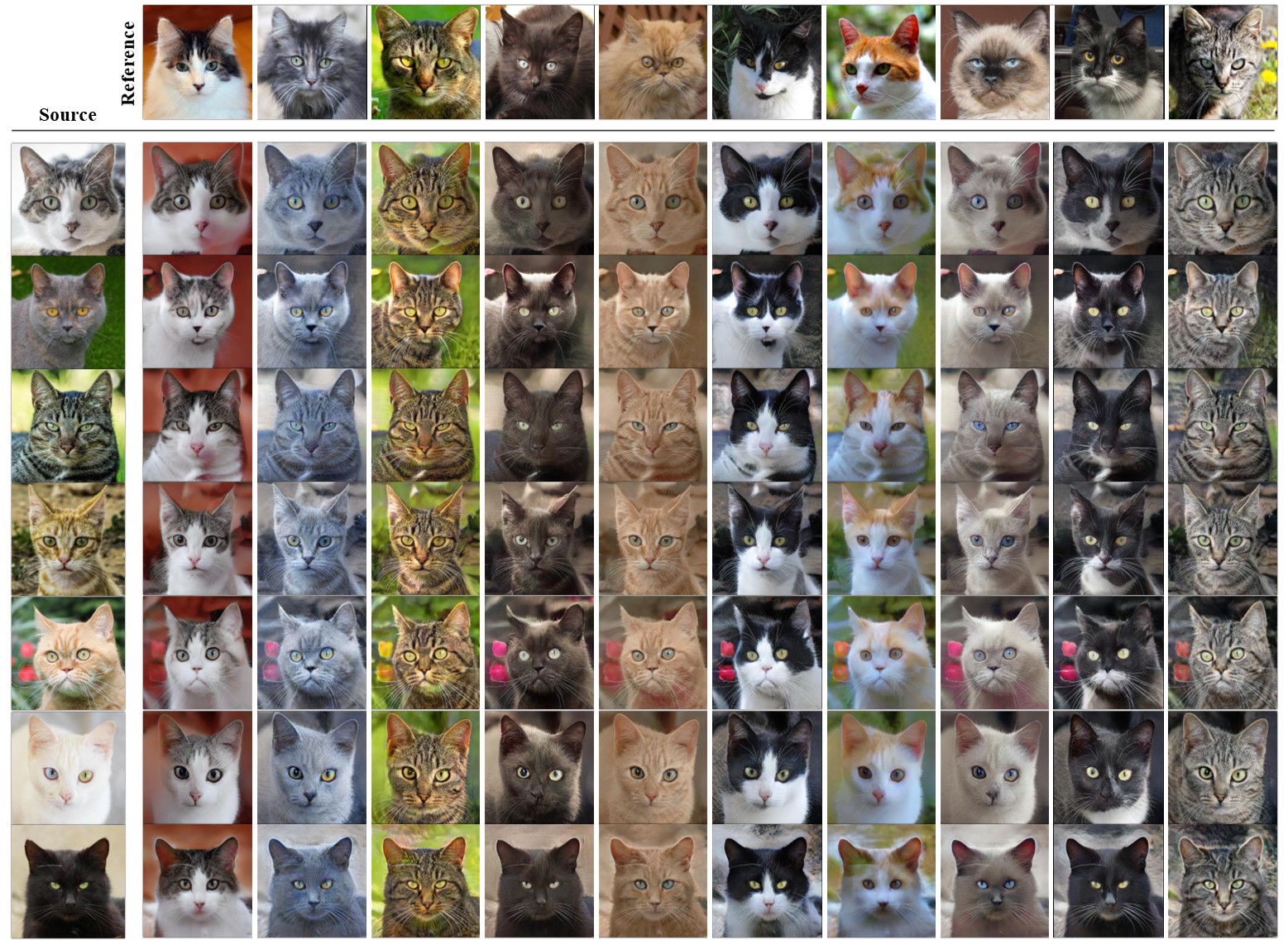}\\
\makebox[1\linewidth][c]{\textbf{(b) Results guided by reference images}}
\vspace{-5mm}
\caption{AFHQ Cat, unsupervised image-to-image translation results.}
\label{fig:AppAFHQCat}
\end{figure*}
}

\newcommand{\figAppAFHQDog}{
\begin{figure*}[b]
\centering
\vspace{-7mm}
\includegraphics[width=0.75\textwidth]{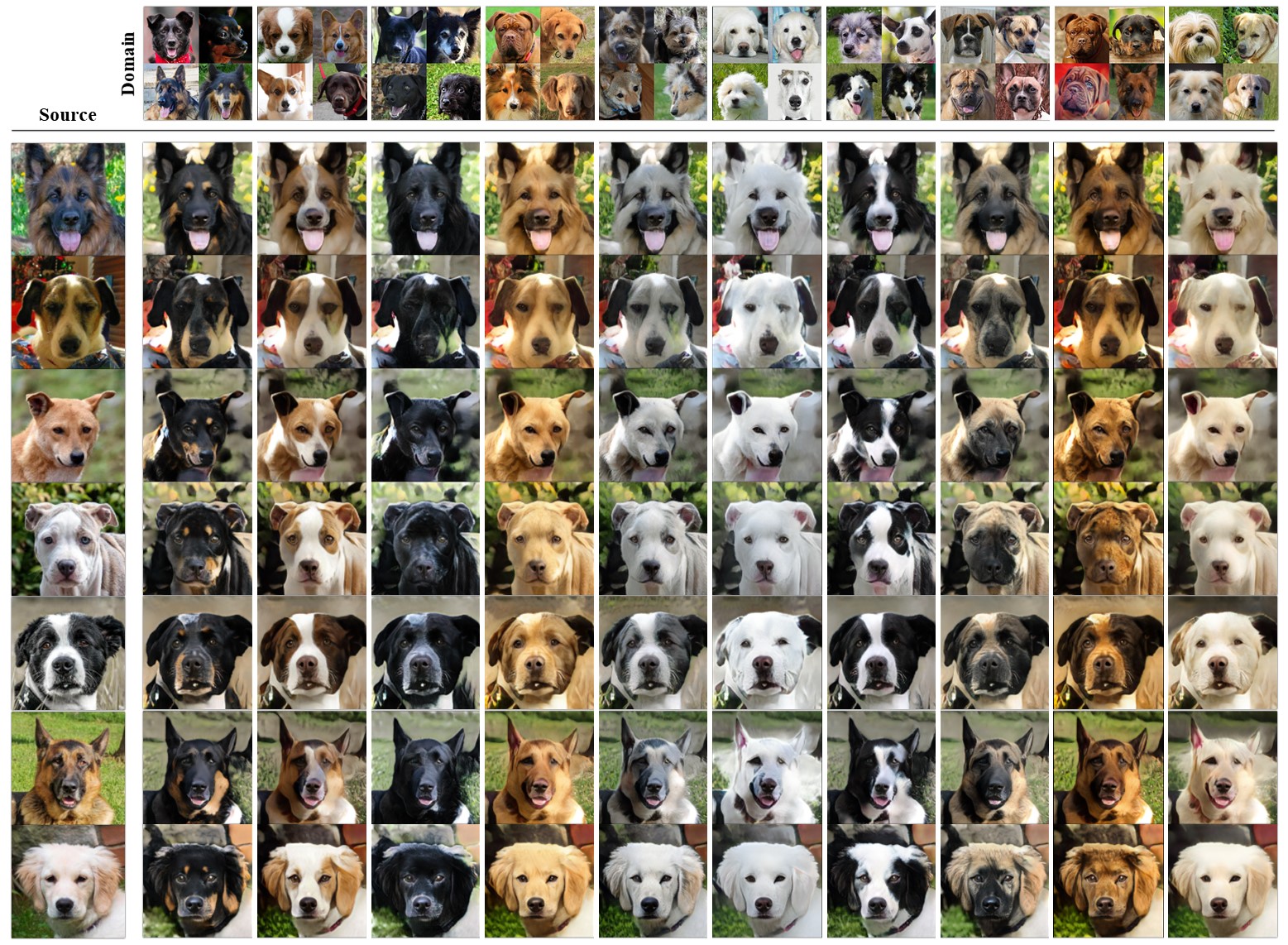}\\
\makebox[1\linewidth][c]{\textbf{(a) Results guided by the average style code of each domain}}\vspace{1mm}
\includegraphics[width=0.75\textwidth]{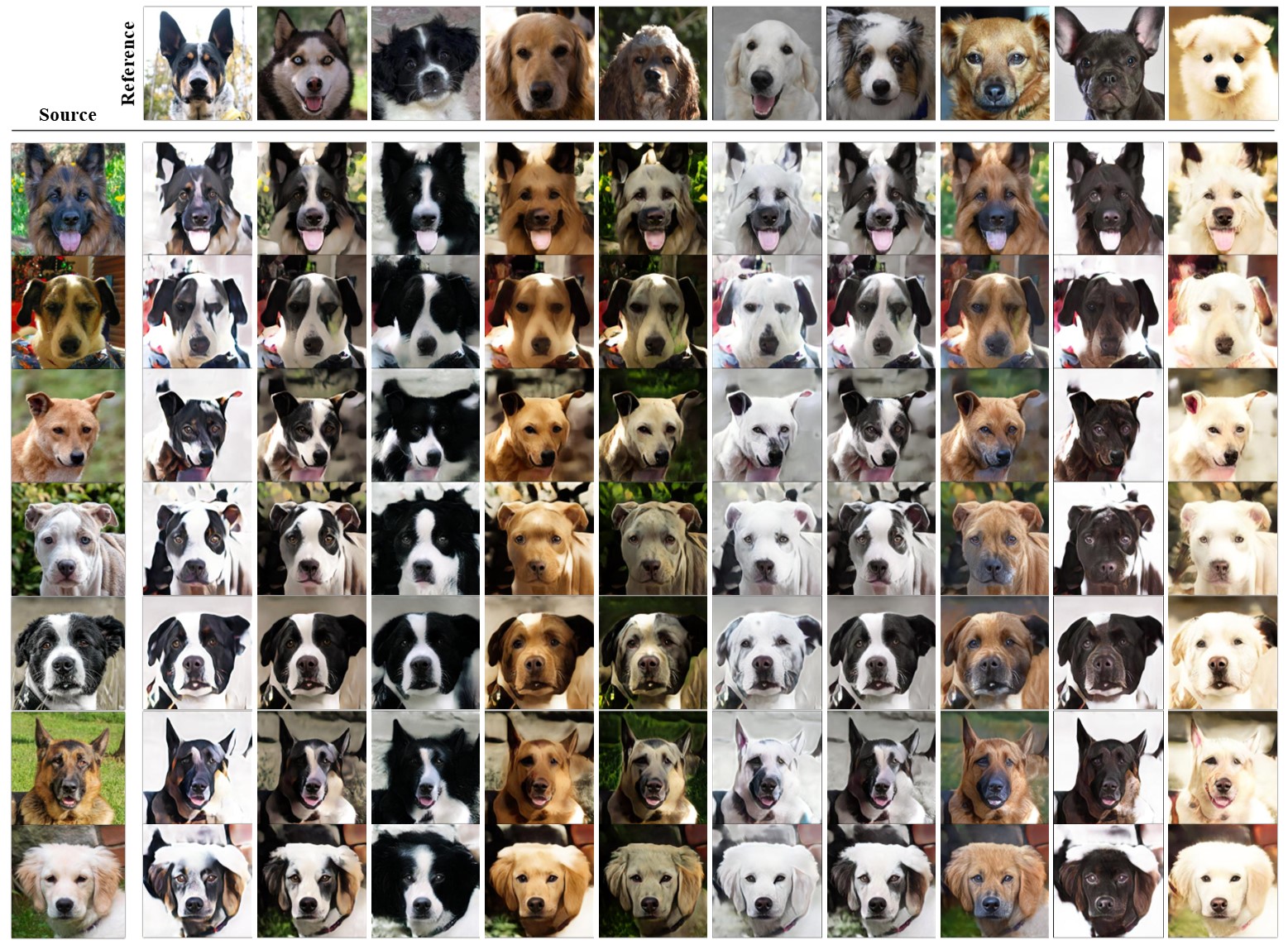}\\
\makebox[1\linewidth][c]{\textbf{(b) Results guided by reference images}}
\vspace{-5mm}
\caption{AFHQ Dogs, unsupervised image-to-image translation results.}
\label{fig:AppAFHQDog}
\end{figure*}
}

\newcommand{\figAppAFHQWild}{
\begin{figure*}[b]
\centering
\vspace{-7mm}
\includegraphics[width=0.75\textwidth]{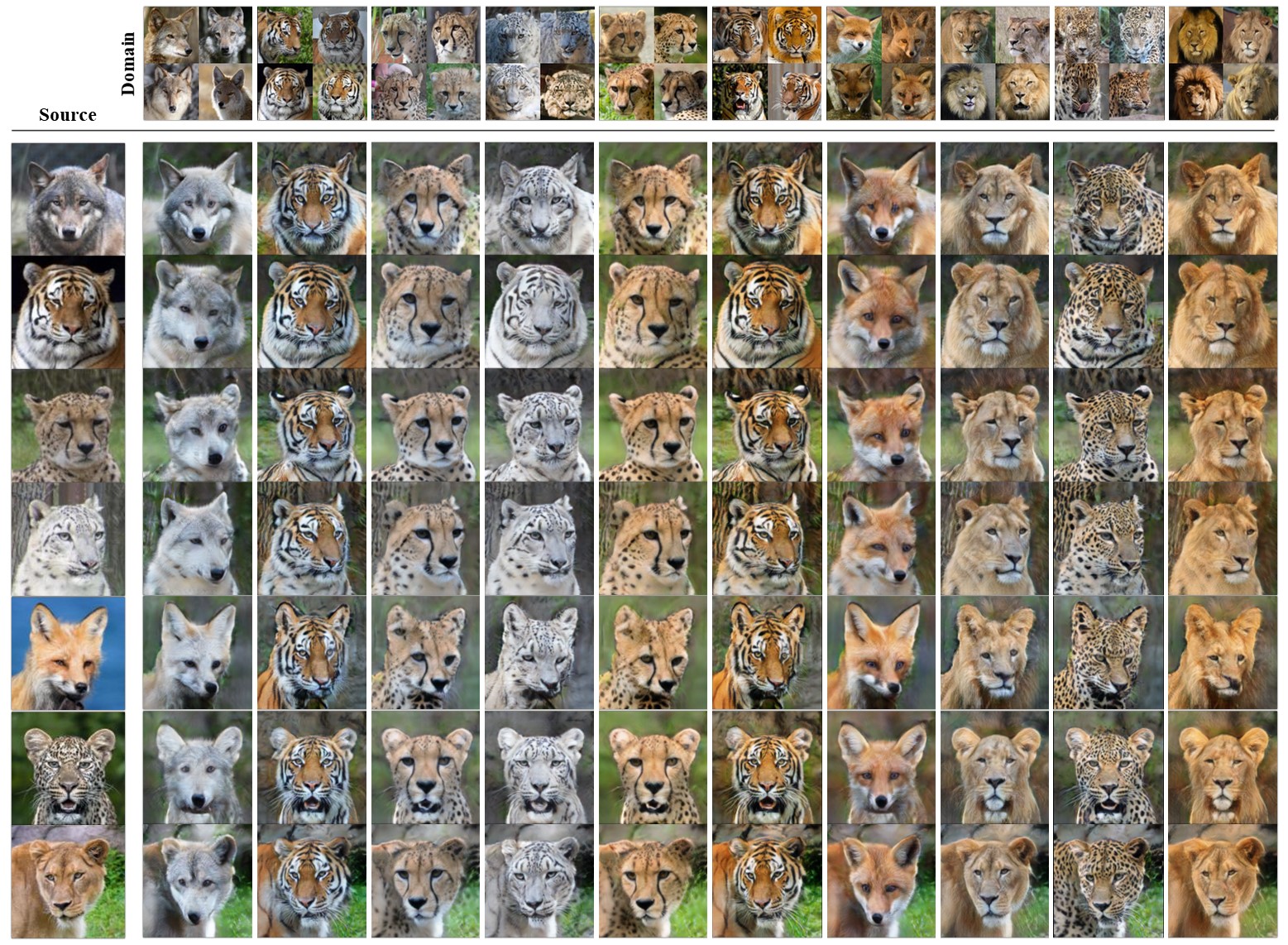}\\
\makebox[1\linewidth][c]{\textbf{(a) Results guided by the average style code of each domain}}\vspace{1mm}
\includegraphics[width=0.75\textwidth]{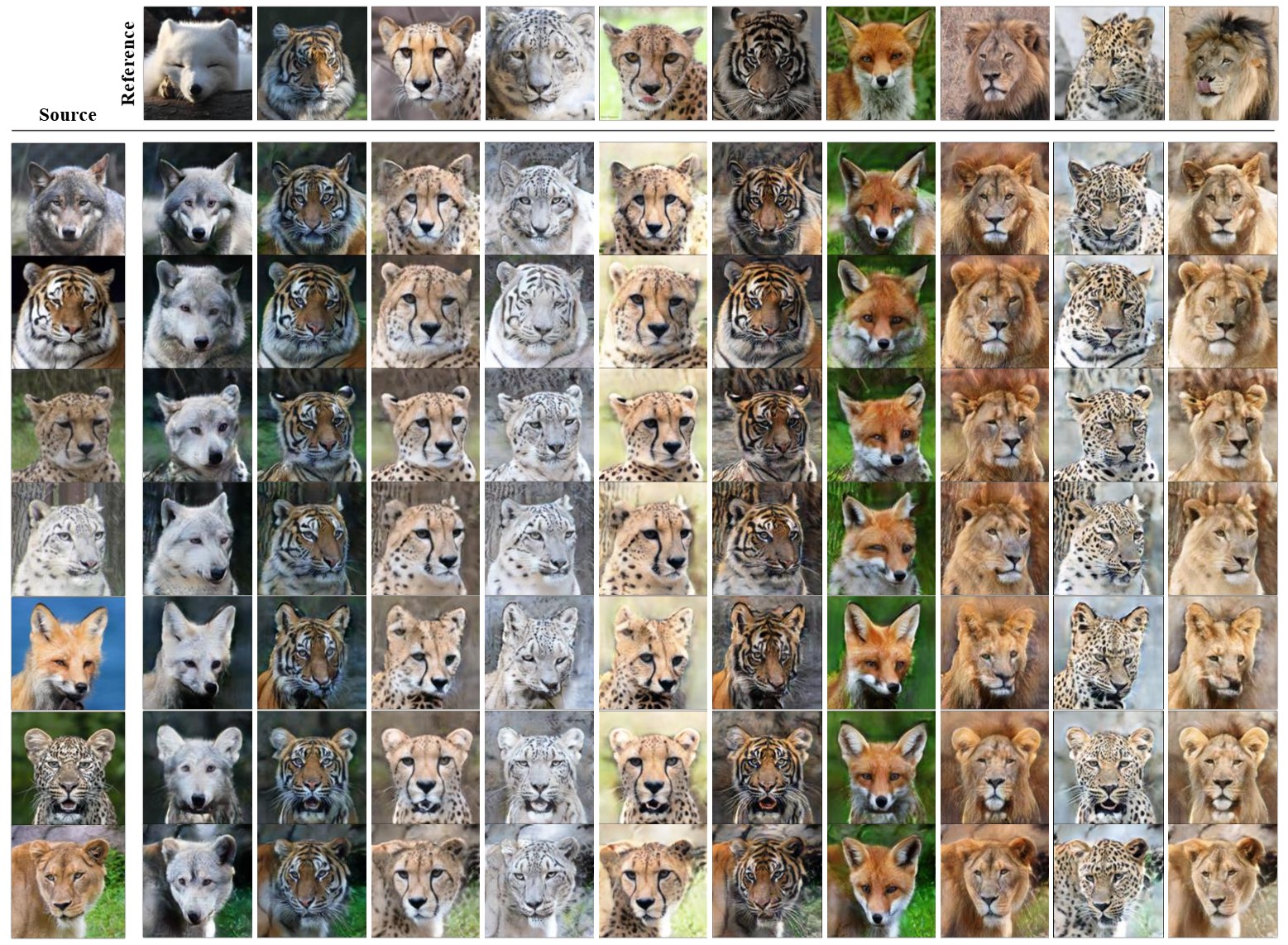}\\
\makebox[1\linewidth][c]{\textbf{(b) Results guided by reference images}}
\vspace{-5mm}
\caption{AFHQ Wild, unsupervised image-to-image translation results.}
\label{fig:AppAFHQWild}
\end{figure*}
}

\newcommand{\figAppFFHQ}{
\begin{figure*}[b]
\centering
\vspace{-7mm}
\includegraphics[width=0.75\textwidth]{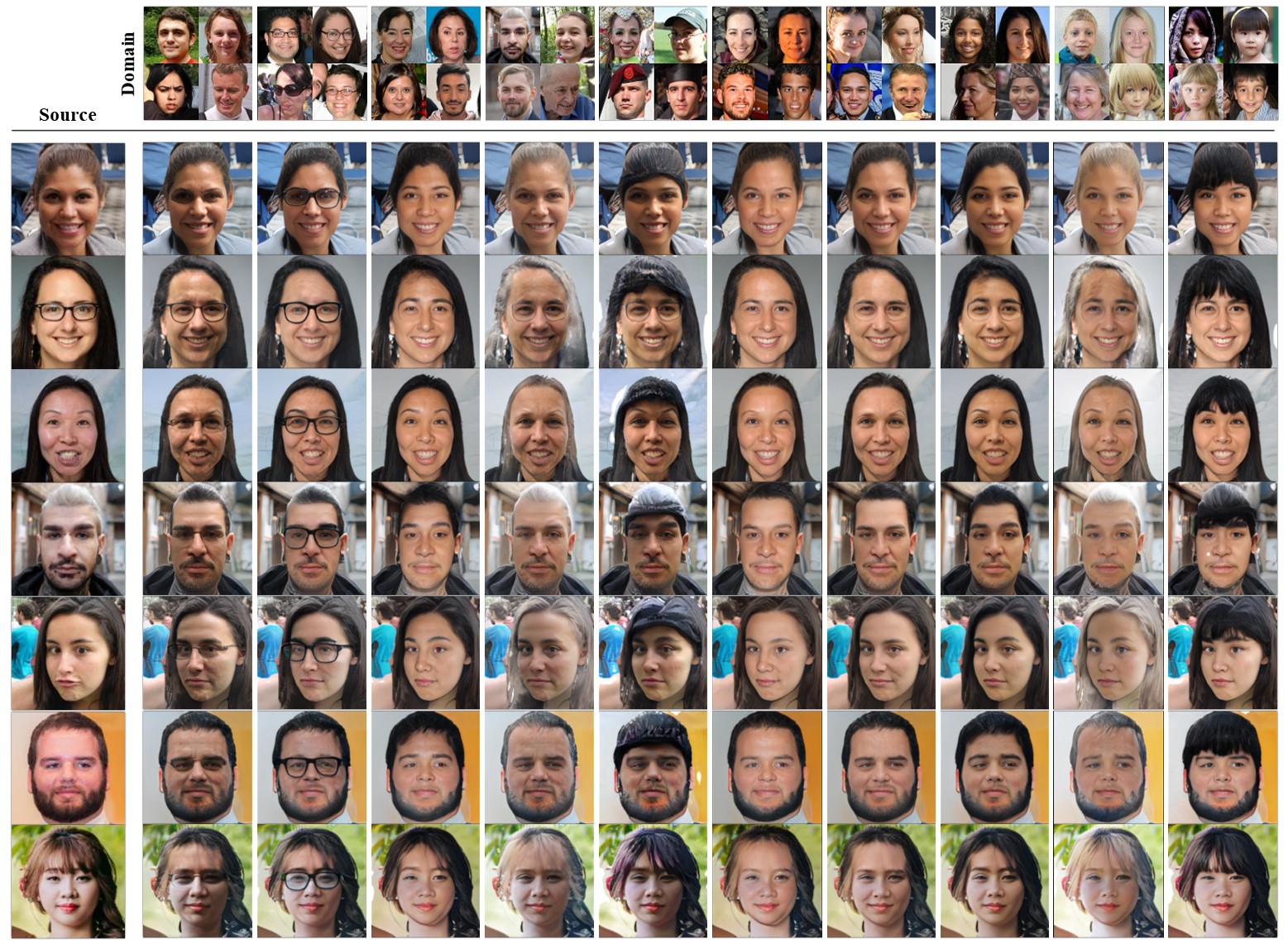}\\
\makebox[1\linewidth][c]{\textbf{(a) Results guided by the average style code of each domain}}\vspace{1mm}
\includegraphics[width=0.75\textwidth]{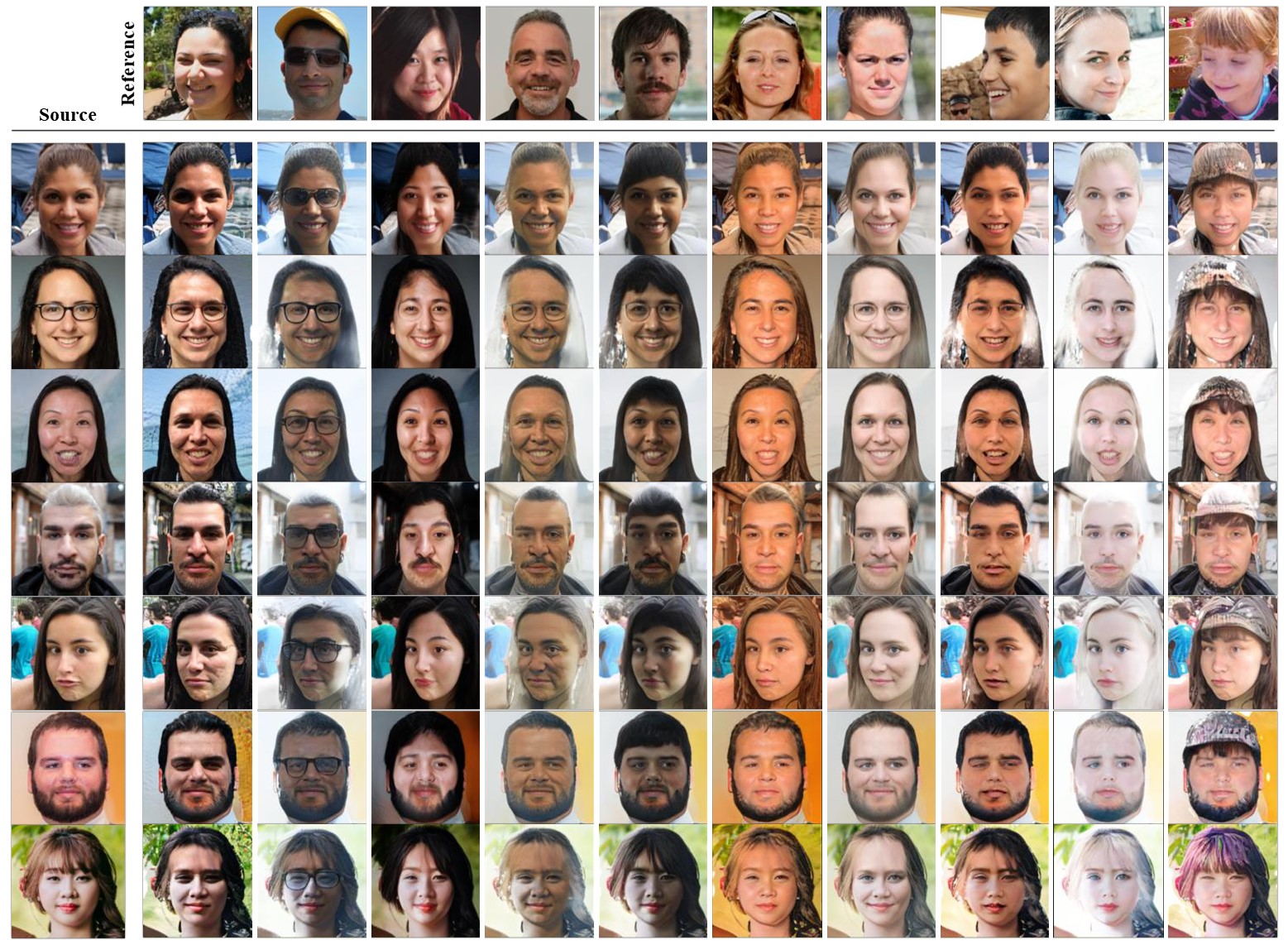}\\
\makebox[1\linewidth][c]{\textbf{(b) Results guided by reference images}}
\vspace{-5mm}
\caption{FFHQ, unsupervised image-to-image translation results.}
\label{fig:AppFFHQ}
\end{figure*}
}

\newcommand{\figAppLSUN}{
\begin{figure*}[b]
\centering
\vspace{-7mm}
\includegraphics[width=0.75\textwidth]{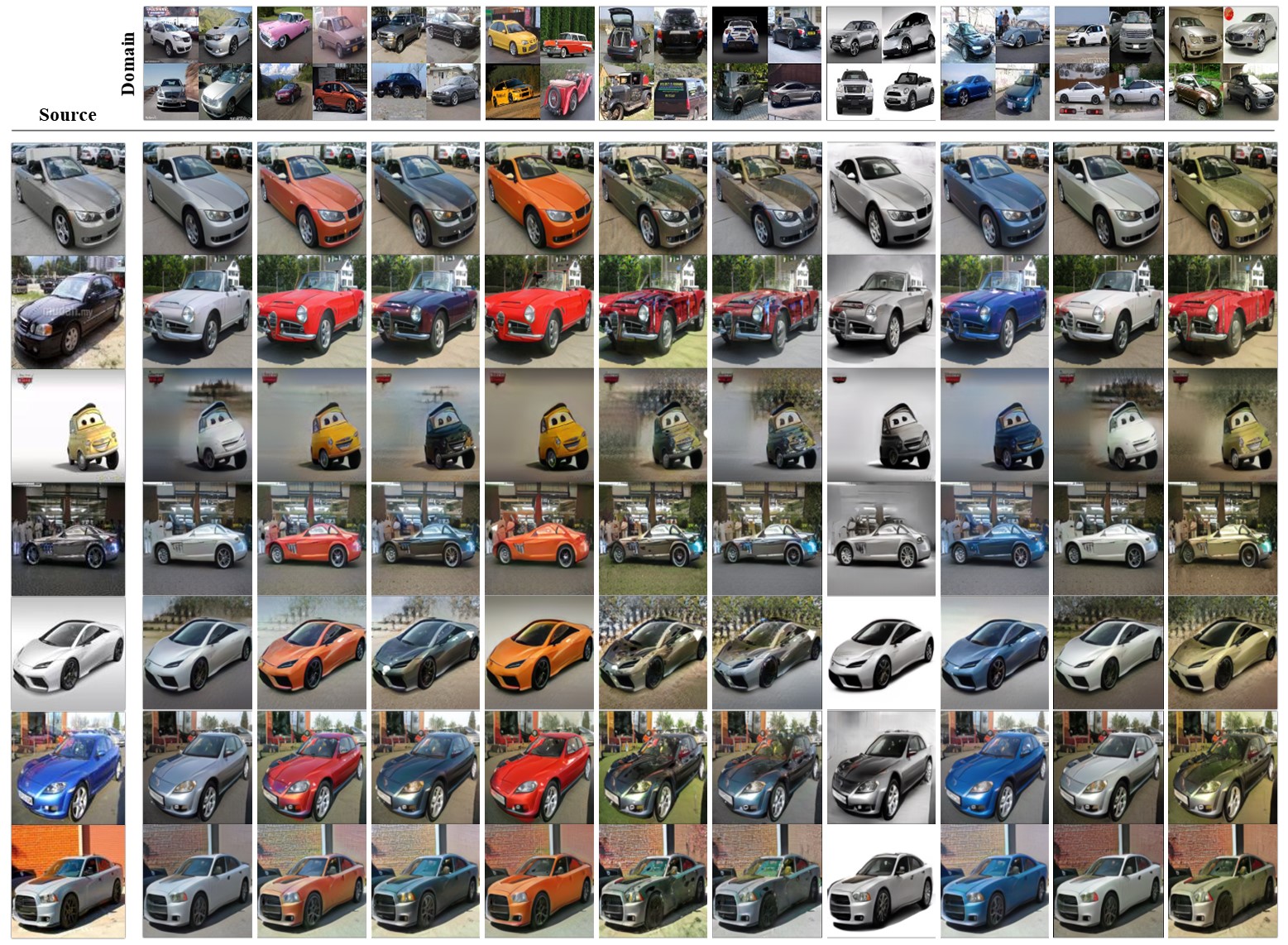}\\
\makebox[1\linewidth][c]{\textbf{(a) Results guided by the average style code of each domain}}\vspace{1mm}
\includegraphics[width=0.75\textwidth]{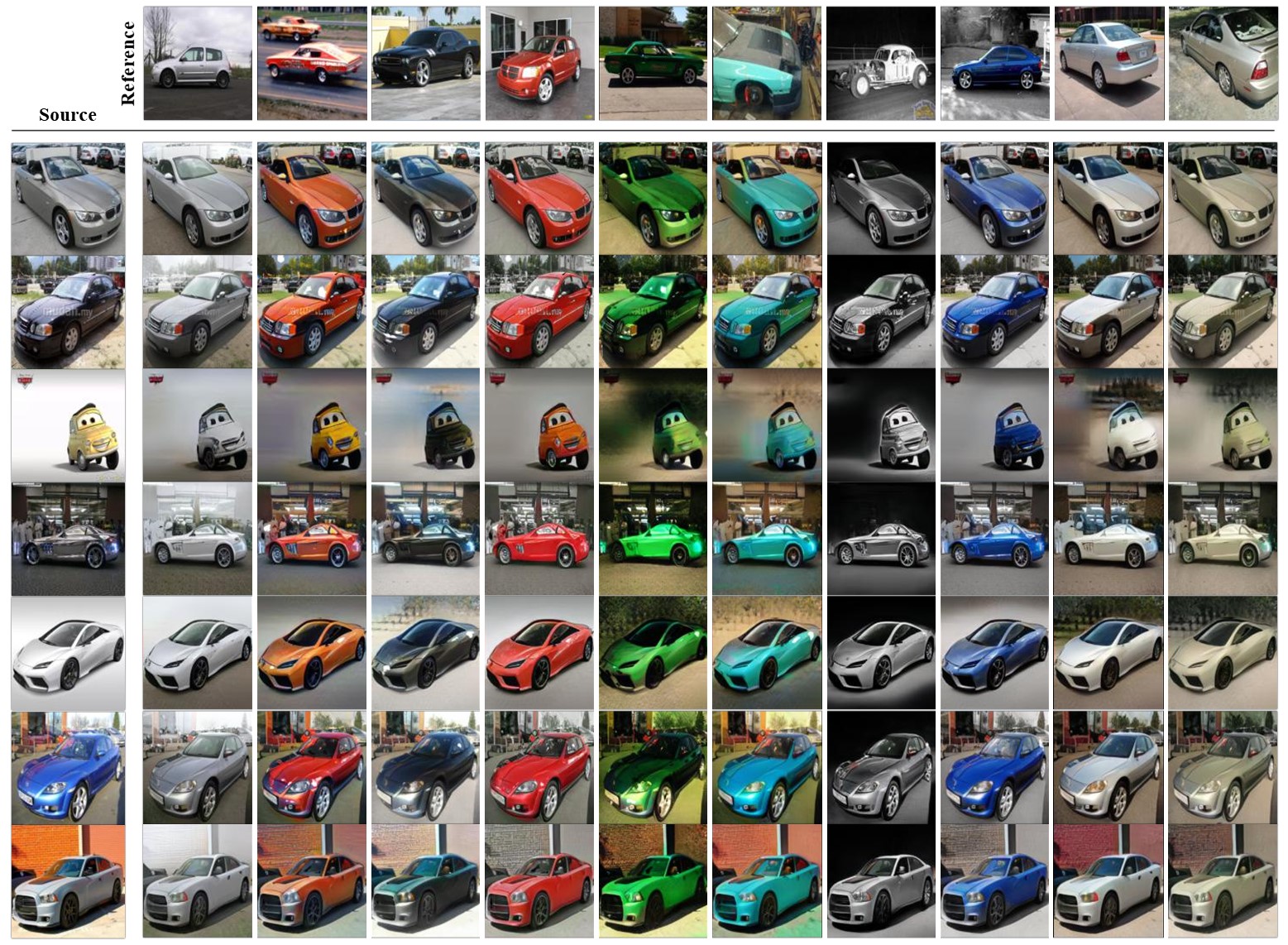}\\
\makebox[1\linewidth][c]{\textbf{(b) Results guided by reference images}}
\vspace{-5mm}
\caption{LSUN Car, unsupervised image-to-image translation results.}
\label{fig:AppLSUN}
\end{figure*}
}

\newcommand{\figAppCatVsFUNIT}{
\begin{figure}[!h]
\centering
\includegraphics[width=0.98\columnwidth]{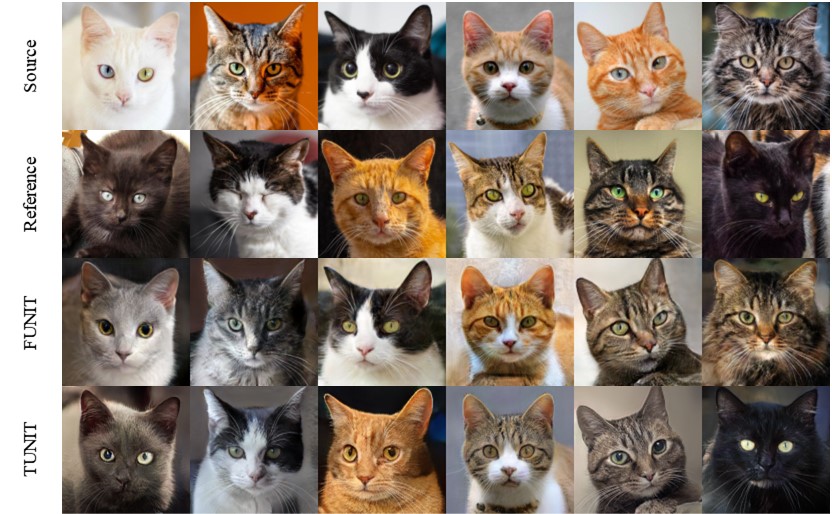}\\
\caption{AFHQ Cat, unsupervised reference-guided image-to-image translation results of FUNIT and TUNIT. The content and the style are from the source and the reference, respectively. While FUNIT usually fails to reflect the style of the reference image, TUNIT generates the fake images with the style -- color, fur texture.}
\label{fig:AppCatVsFUNIT}
\end{figure}
}

\newcommand{\figAppWildVsFUNIT}{
\begin{figure}[!h]
\centering
\includegraphics[width=0.98\columnwidth]{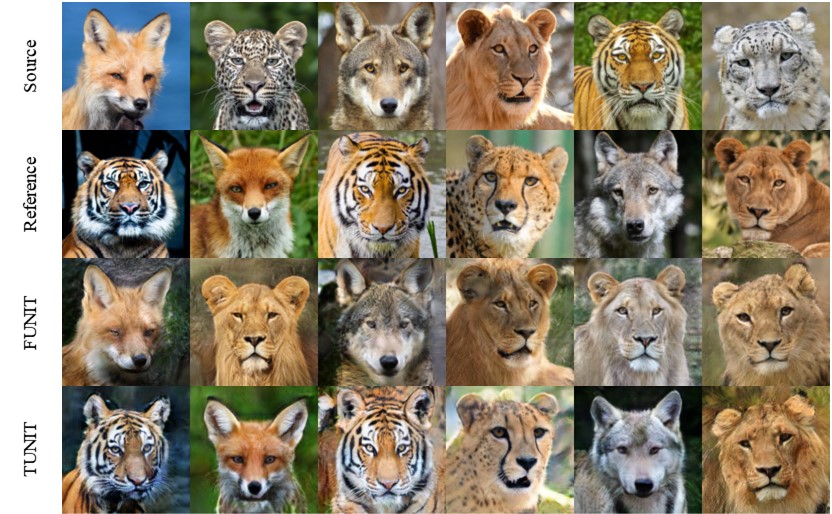}\\
\caption{AFHQ Wild, unsupervised reference-guided image-to-image translation results of FUNIT and TUNIT. FUNIT rarely reflects the correct style of the reference image -- the species, on the other hand, TUNIT translates the source image to the correct species.}
\label{fig:AppWildVsFUNIT}
\end{figure}
}

\newcommand{\figAppLSUNVsFUNIT}{
\begin{figure}[!h]
\centering
\vspace{1.555cm}
\includegraphics[width=0.98\columnwidth]{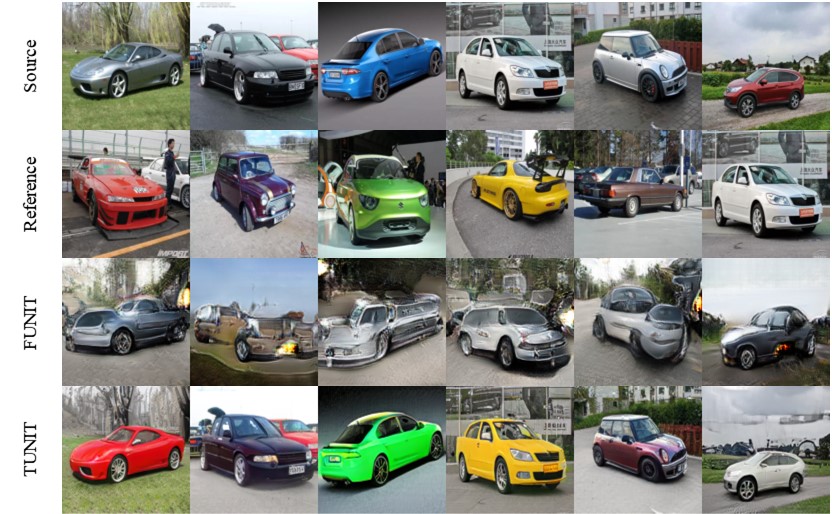}\\
\caption{LSUN Car, unsupervised reference-guided image-to-image translation results of FUNIT and TUNIT. While TUNIT generates plausible and changes the color of the source image to that of the reference image, FUNIT not also generates unrealistic image but also fails to changes the color.}
\label{fig:AppLSUNVsFUNIT}
\end{figure}
}

\newcommand{\figAppFFHQVsFUNIT}{
\begin{figure}[!h]
\centering
\includegraphics[width=0.98\columnwidth]{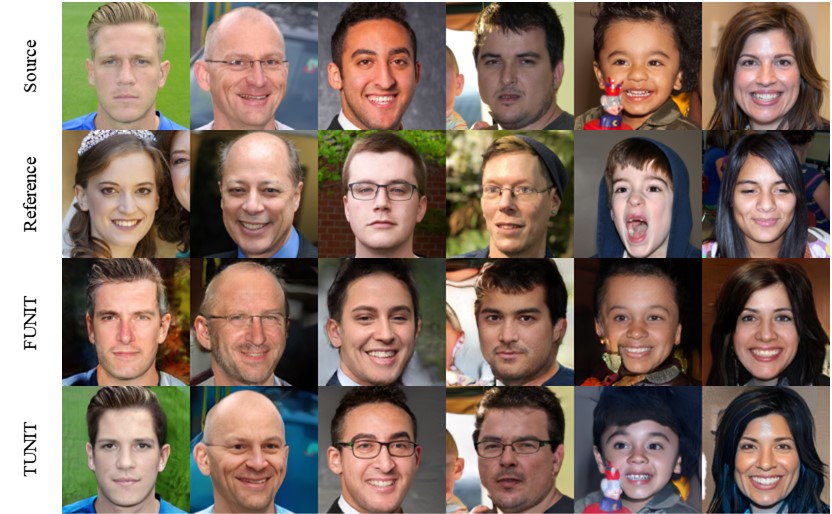}\\
\caption{FFHQ, unsupervised reference-guided image-to-image translation results of FUNIT and TUNIT. Our model, TUNIT can remove or add the glasses to the source while preserving the identity better than FUNIT. In addition, TUNIT can change the hair color (last column) and the hair style -- especially, bang (fifth column). It is hard to specify the definition of domains in the results of FUNIT while domains of TUNIT are more interpretable.}
\label{fig:AppFFHQVsFUNIT}
\end{figure}
}

\newcommand{\figAppStoW}{
\begin{figure*}[b]
\centering
\vspace{-7mm} 
\includegraphics[width=0.85\textwidth]{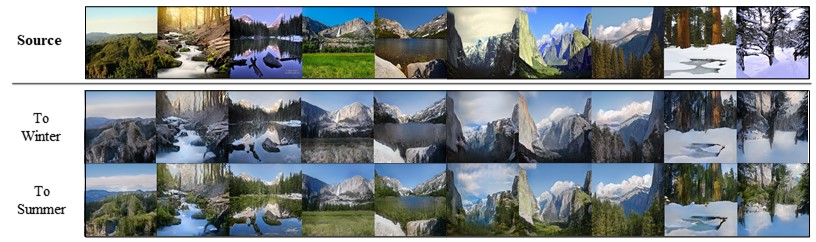}\\
\makebox[1\linewidth][c]{\textbf{(a) Results guided by the average style code of each domain}}\vspace{1mm}
\includegraphics[width=0.85\textwidth]{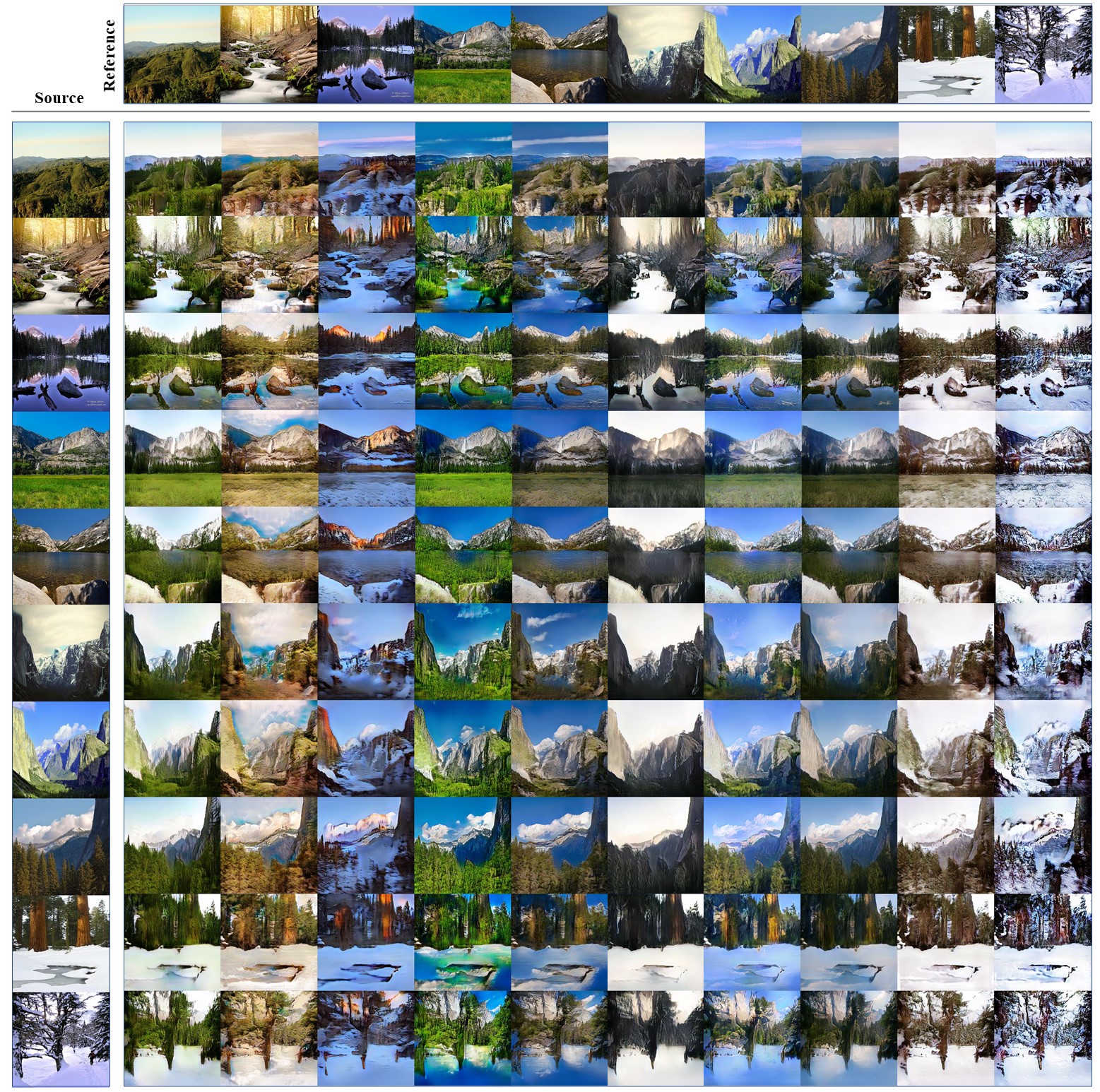}\\
\makebox[1\linewidth][c]{\textbf{(b) Results guided by reference images}}
\vspace{-5mm}
\caption{Summer2Winter (S2W), unsupervised image-to-image translation results.}
\label{fig:AppStoW} 
\end{figure*}
}

\newcommand{\figAppPtoU}{
\begin{figure*}[b]
\centering
\vspace{-7mm} 
\includegraphics[width=0.85\textwidth]{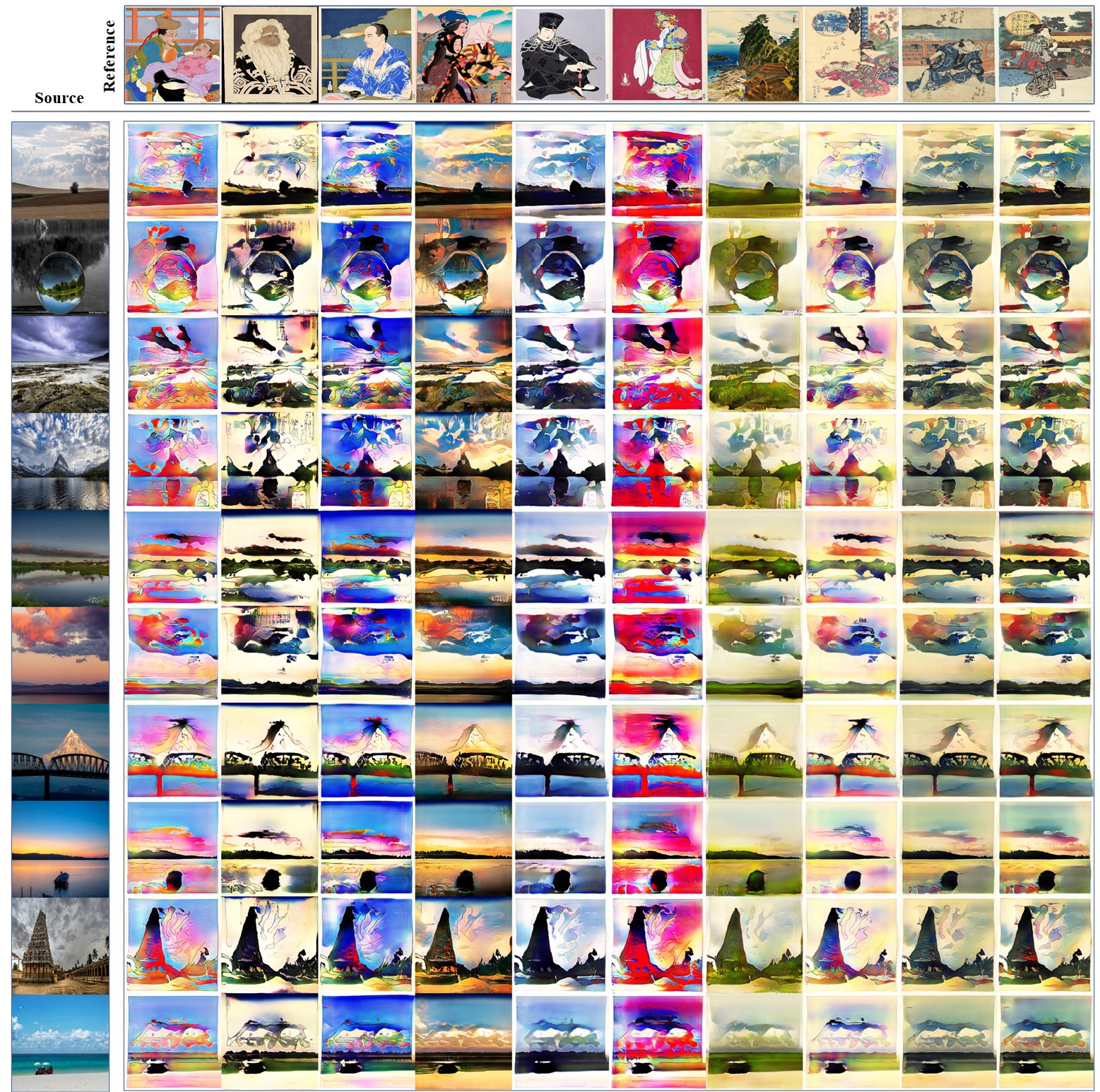}\\
\makebox[1\linewidth][c]{\textbf{Results guided by reference images}}
\vspace{-5mm}
\caption{Photo2Ukiyoe, unsupervised image-to-image translation results.}
\label{fig:AppPtoU} 
\end{figure*}
}

\newcommand{\figAppSEMIT}{
\begin{figure*}[b]
\centering
\vspace{-7mm} 
\includegraphics[width=0.85\textwidth]{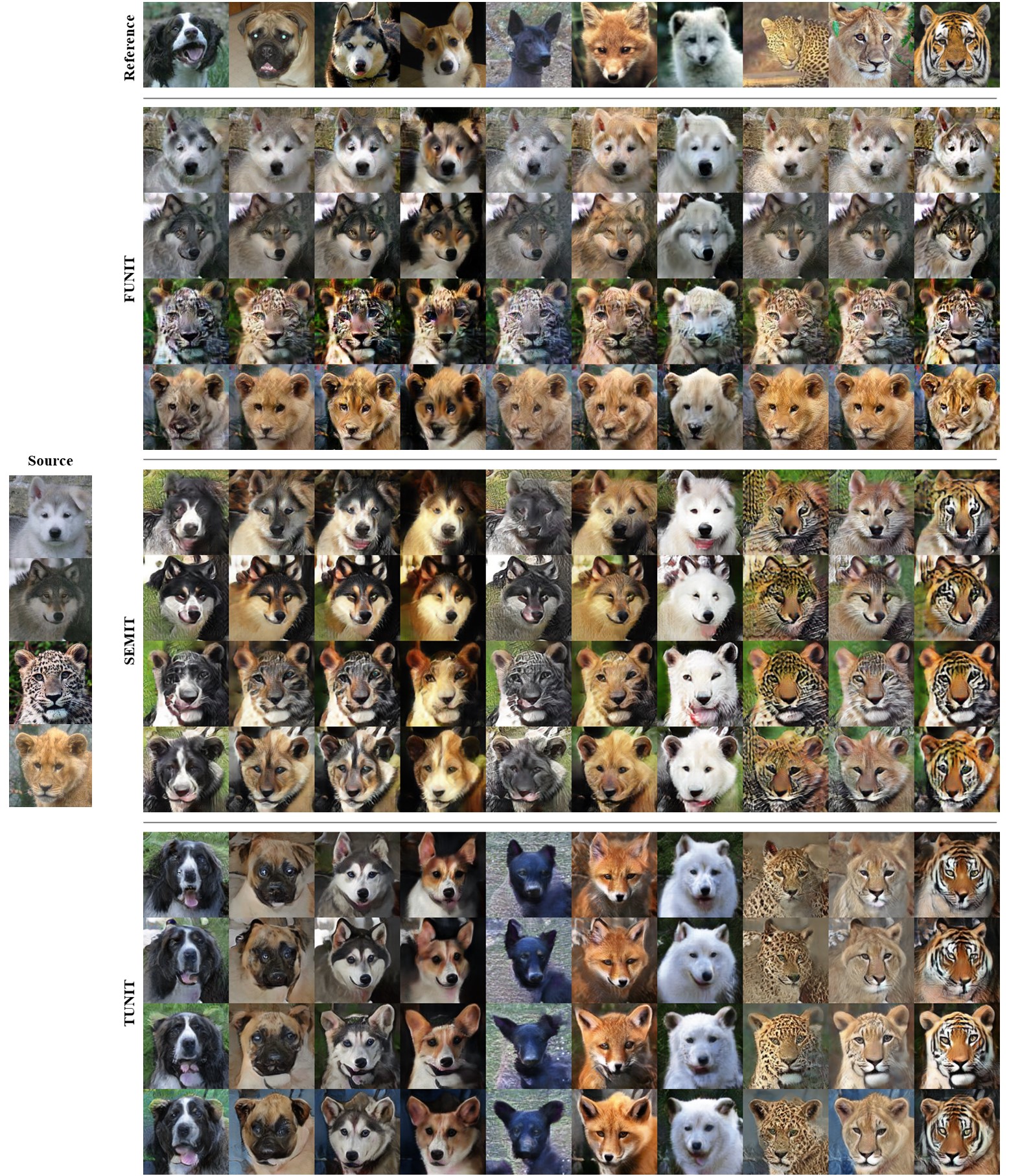}\\
\caption{Qualitative comparison with SEMIT and FUNIT, semi-supervised translation results on AnimalFaces-149 (1\% of labeled samples are used).}
\label{fig:AppSEMIT} 
\end{figure*}
}

\newcommand{\figAppLPIPS}{
\begin{figure*}[b]
\centering
\includegraphics[width=1\textwidth]{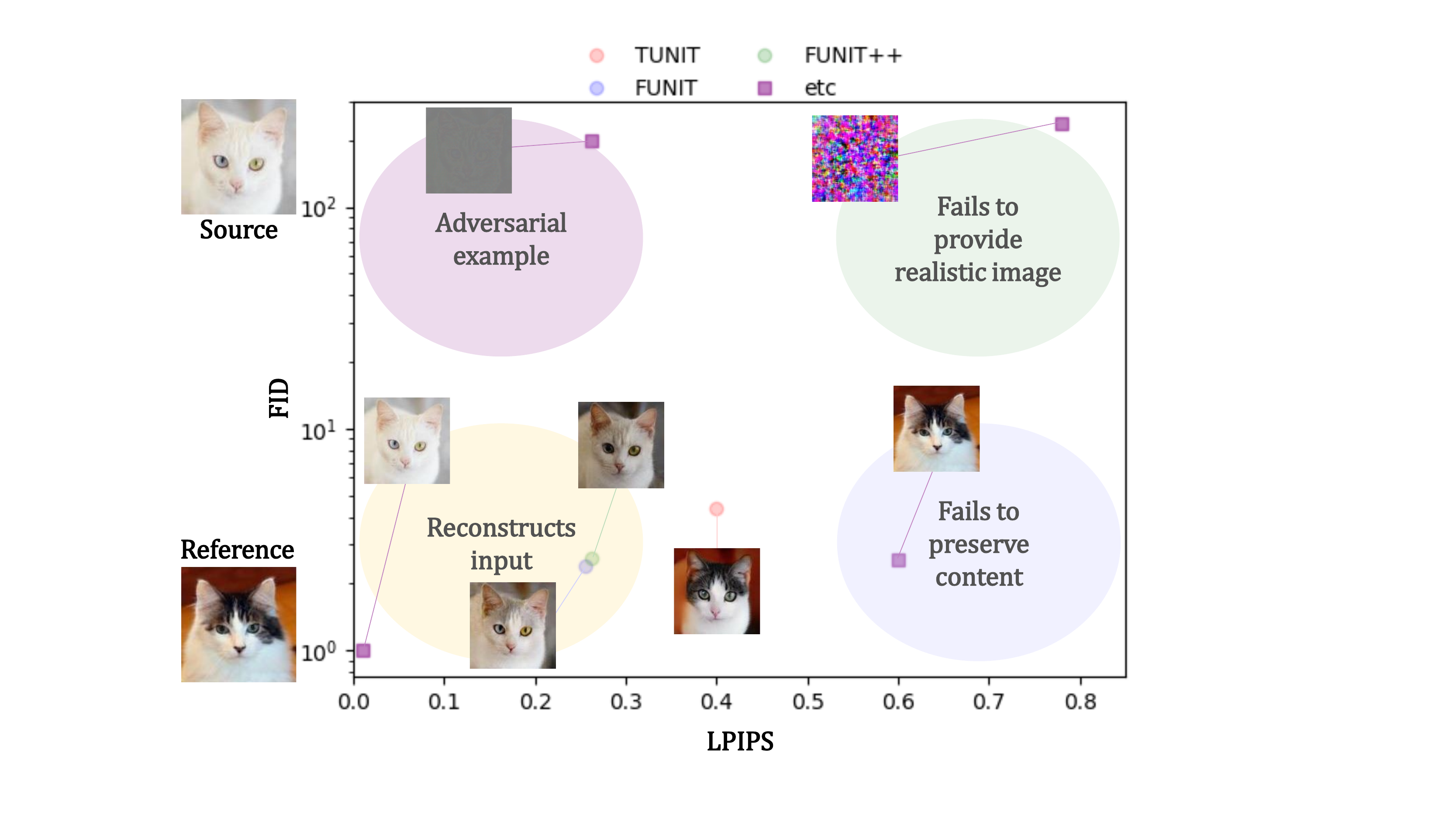}\\
\vspace{-5mm}
\caption{\textbf{LPIPS and FID of models and their status.}}
\label{fig:AppLPIPS} 
\end{figure*}
}

%% file: table.tex
\newcommand{\tabmain}{
\begin{table*}[t]
\centering
\newcolumntype{x}{>{\centering\arraybackslash\hspace{0pt}}p{17mm}}
\newcolumntype{s}{>{\centering\arraybackslash\hspace{0pt}}p{6mm}}
\small{
\begin{tabular}{|l@{\hspace{2.5mm}}l|p{6mm}p{17mm}p{6mm}|p{6mm}p{17mm}p{6mm}|}
\hline
& \multirow{2}{*}{\textbf{Configuration}}
    & \multicolumn{3}{c|}{\textbf{AnimalFaces-10}}
    & \multicolumn{3}{c|}{\textbf{Food-10}}
\\
&
    & mFID & \makebox[17mm][c]{D \& C} & Acc.
    & mFID & \makebox[17mm][c]{D \& C} & Acc.
\\ \hline
\\[-0.8em]
\arch{a} & Baseline FUNIT  (supervised)\ \                   %
    & 74.0          & 0.749 / 0.671   & \textbf{1.000}         %
    & 68.4          & 0.989 / 0.782   & \textbf{1.000}         %
\\[0.3mm]
\arch{b} & (\arch{a}) + Improved G \& D (supervised) \ \                                 %
    & \textbf{46.2}          & \textbf{0.896}  / \textbf{0.732}         & \textbf{1.000}         %
    & \textbf{57.6}          & \textbf{1.284} / \textbf{0.857}         & \textbf{1.000}         %
\\[0.3mm] \hline
\\[-0.8em]
\arch{c} & (\arch{b}) + K-means on image space\ \                                 %
    & 110.7         & 0.822  / 0.615         & 0.215         %
    & 90.7          & 0.849 / 0.648         & 0.201         %
\\[0.3mm]
\arch{d} & (\arch{b}) + K-means on feature space\ \                          %
    & 76.2          & 0.770 / 0.597         & 0.428         %
    & 64.6          & 0.968 / 0.808         & 0.331         %
\\[0.3mm]
\arch{e} & (\arch{b}) + Differentiable clustering\ \                        %
    & 73.5          & 0.940 / 0.588         & 0.680         %
    & 64.2          & 1.038 / 0.819         & 0.542         %
\\[0.3mm]
\arch{f} & TUNIT w/ sequential training\ \                                      %
    & \textbf{46.0}    & 1.060 / 0.789         & \textbf{0.850}   %
    & 61.1    & 0.908 / 0.777         & \textbf{0.860}   %
\\[0.3mm]
\arch{g} & TUNIT w/ joint training\ \                                      %
    & 47.7    & \textbf{1.039} / \textbf{0.805}         & 0.841   %
    & \textbf{52.2}    & \textbf{1.079} / \textbf{0.875}         & 0.848   %
\\[0.3mm] \hline
\end{tabular}\vspace{0mm}}
\caption{\small{
Comparison results. mFID, Density / Coverage (D\&C), and classification accuracy (Acc) of each training configuration. Note that the configurations (\arch{a}) - (\arch{b}) use ground-truth class labels, while (\arch{c}) - (\arch{g}) use pseudo-labels. We \textbf{bold} the best results separately for supervised and unsupervised settings. For D\& C, we boldify the one which has the best coverage that has a clear maximum.}
}\vspace{-3mm}
\label{tab:main}
\end{table*}
}

\newcommand{\tabKcomparisonn}{
\begin{table}[t]
\centering
\begin{minipage}[]{0.4\columnwidth}
    \centering
    \includegraphics[keepaspectratio=false, width=1.0\linewidth,height=1.0\linewidth]{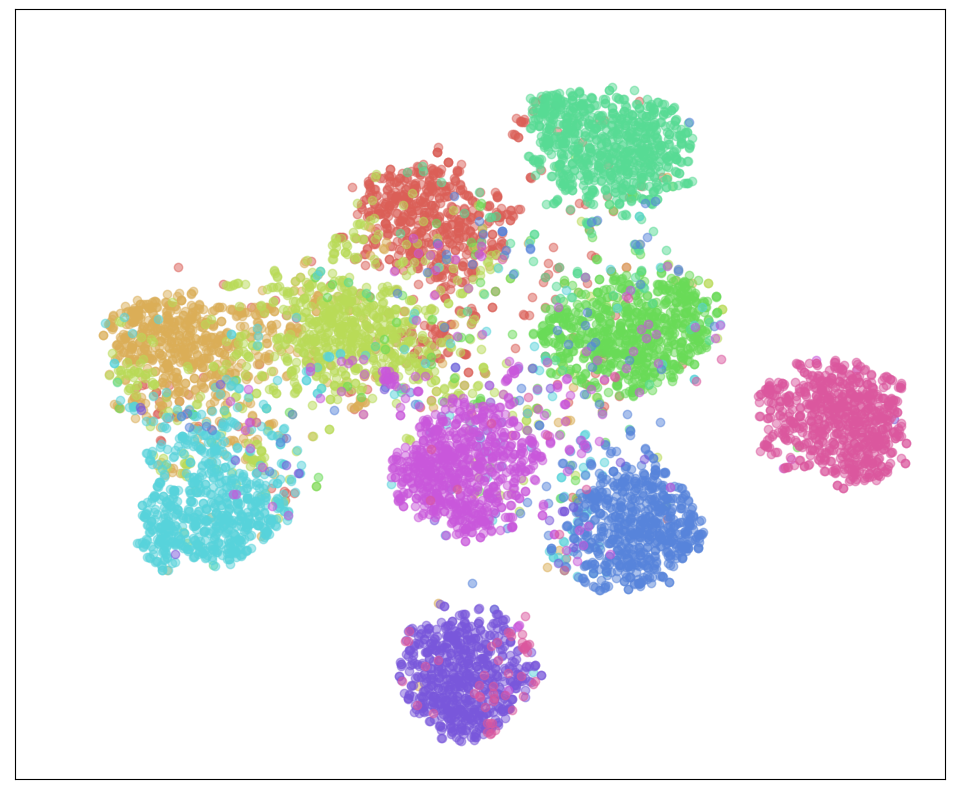}
    \footnotesize{(a) t-SNE of $\hat{K}$=10}
\end{minipage}\quad
\begin{minipage}[]{0.4\columnwidth}
    \centering
    \includegraphics[keepaspectratio=false, width=1.0\linewidth,height=1.0\linewidth]{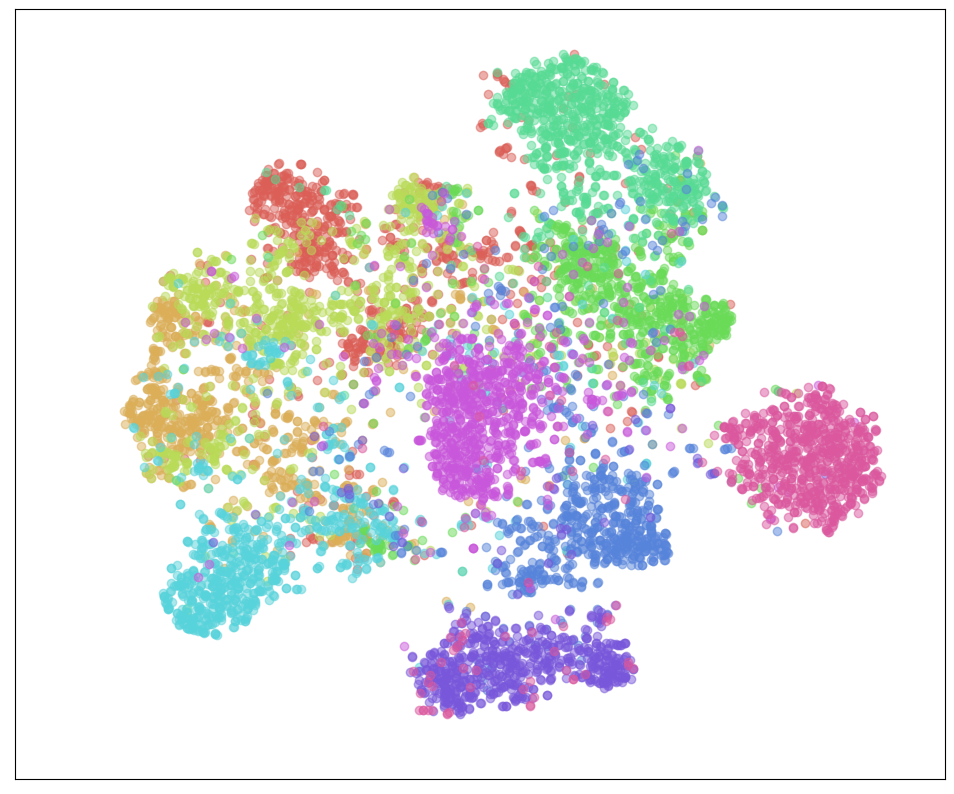}
    \footnotesize{(b) t-SNE of $\hat{K}$=20}
\end{minipage}\quad

\begin{minipage}[]{0.9\columnwidth}
    \centering
    \footnotesize{
    \begin{tabular}{|l@{\hspace{1.0mm}}l|p{5mm}p{17mm}|p{5mm}p{17mm}|}
    \hline
    & \multirow{2}{*}{\textbf{$\hat{K}$}}
        & \multicolumn{2}{c|}{\textbf{AnimalFaces-10}}
        & \multicolumn{2}{c|}{\textbf{Food-10}}
    \\ 
    &
        & \makebox[6mm][c]{mFID} & \makebox[17mm][c]{D \& C}
        & \makebox[6mm][c]{mFID} & \makebox[17mm][c]{D \& C}
    \\ \hline
    \\[-0.8em]
    & \footnotesize{$1$} \ \              
        & 129.6          & 0.561 / 0.512        
        & 95.1              & 1.113 / 0.771        
    \\[0.3mm]
    & \footnotesize{$4$} \ \ 
        & 77.7          & 0.879 / 0.738        
        & 67.4             & 0.851 / 0.785        
    \\[0.3mm]
    & \footnotesize{$7$} \ \  
        & 62.7         & 1.016 / 0.729
        & 52.7            & 1.024 / 0.846 
    \\[0.3mm]
    & \footnotesize{$10$} \ \  
        & \textbf{47.7}          & \textbf{1.039} / \textbf{0.805} 
        & \textbf{52.2}             & \textbf{1.079} / \textbf{0.875}         %
    \\[0.3mm]
    & \footnotesize{$13$}\ \   
        & 56.8          & 0.993 / 0.720  
        & 54.8             & 0.970 / 0.845         
    \\[0.3mm]
    & \footnotesize{$16$}\ \  
        & 54.1           & 1.093 / 0.782   
        & 54.8              & 1.029 / 0.857       
    \\[0.3mm]
    & \footnotesize{$20$}\ \  
        & 55.4           & 1.019 / 0.778   
        & 57.7              & 0.937 / 0.846     
    \\[0.3mm] 
     & \footnotesize{$50$}\ \  
    & 63.8           & 0.858 / 0.701
    & 60.8              & 1.067 / 0.837
\\[0.3mm]
 & \footnotesize{$500$}\ \  
    & 67.2           & 0.921 / 0.694
    & 63.2              & 0.986 / 0.826
\\[0.3mm]
 & \footnotesize{$1000$}\ \  
    & 66.9           & 0.908 / 0.707
    & 60.7              & 0.945 / 0.845
\\[0.3mm]
    
    \hline
\end{tabular}}
\end{minipage}
\caption{
\small{t-SNE visualization of the model with (a) $\hat{K}$=10 and (b) $\hat{K}$=20 trained on AnimalFaces-10 and quantitative evaluation of our method by varying the number of pseudo domains $\hat{K}$. Each point is colored with the ground-truth labels. As shown in t-SNE visualizations, even if $\hat{K}$ is set to overly larger than the actual number of domains, the guiding network clusters the domains reasonably well. For D\& C, we bold the one which has the best coverage that has a clear maximum.}
}
\vspace{-5mm}
\label{tab:k_comparison}
\end{table}
}

\newcommand{\tabsemi}{
\newcommand{\hrow}{5mm}
\begin{table*}[t]
\centering
\newcolumntype{x}{>{\centering\arraybackslash\hspace{-2mm}}p{\hrow}}
\footnotesize{
\begin{tabular}{|l@{\hspace{1mm}}l|xxxx|xxxx|xxxx|xxxx|}
\hline
& \multirow{2}{*}{\textbf{Model}}
    & \multicolumn{8}{c|}{\textbf{AnimalFaces-10}}
    & \multicolumn{8}{c|}{\textbf{Food-10}}
\\
&
 & \makebox[\hrow][c]{1\%} & \makebox[\hrow][c]{2\%} & \makebox[\hrow][c]{4\%} & \makebox[\hrow][c]{8\%}
 & \makebox[\hrow][c]{20\%} & \makebox[\hrow][c]{40\%} & \makebox[\hrow][c]{60\%} & \makebox[\hrow][c]{80\%}
 & \makebox[\hrow][c]{1\%} & \makebox[\hrow][c]{2\%} & \makebox[\hrow][c]{4\%} & \makebox[\hrow][c]{8\%}
 & \makebox[\hrow][c]{20\%} & \makebox[\hrow][c]{40\%} & \makebox[\hrow][c]{60\%} & \makebox[\hrow][c]{80\%}
\\ \hline
\\[-0.8em]
& FUNIT~\cite{2019FUNIT} \ \                   %
    & 179.8    & 174.3     & 154.3   & 144.0      %
    & 124.4    & 106.4     & 96.0   & 79.6      %
    & 195.7     & 159.2      & 141.2  & 135.2               %
    & 111.4     & 85.8      & 74.8  & 70.3               %
\\[0.3mm]
& SEMIT~\cite{wang2020semit} \ \                   %
    & 68.9    & 63.0     & 63.5   & 60.3      %
    & 57.3    & 64.3     & 66.1   & 62.8      %
    & 70.7     & 66.7      & 65.3  & 65.1               %
    & 65.2     & 62.8      & 61.6  & 64.1               %
\\[0.3mm]
& TUNIT (ours)\ \                        
    & \textbf{42.0}    & \textbf{42.6}     & \textbf{43.9}   & \textbf{46.2}               %
    & \textbf{42.0}    & \textbf{42.6}     & \textbf{43.9}   & \textbf{46.2}               %
    & \textbf{53.6}     & \textbf{56.2}      & \textbf{52.8} & \textbf{53.4}                %
    & \textbf{53.6}     & \textbf{56.2}      & \textbf{52.8} & \textbf{53.4}                %
\\[0.3mm] \hline

& \multirow{2}{*}{\textbf{Model}}
    & \multicolumn{8}{c|}{\textbf{AnimalFaces-149}}
    & \multicolumn{8}{c|}{\textbf{Food-101}}
\\
&
 & \makebox[\hrow][c]{1\%} & \makebox[\hrow][c]{2\%} & \makebox[\hrow][c]{4\%} & \makebox[\hrow][c]{8\%}
 & \makebox[\hrow][c]{20\%} & \makebox[\hrow][c]{40\%} & \makebox[\hrow][c]{60\%} & \makebox[\hrow][c]{80\%}
 & \makebox[\hrow][c]{1\%} & \makebox[\hrow][c]{2\%} & \makebox[\hrow][c]{4\%} & \makebox[\hrow][c]{8\%}
 & \makebox[\hrow][c]{20\%} & \makebox[\hrow][c]{40\%} & \makebox[\hrow][c]{60\%} & \makebox[\hrow][c]{80\%}
\\ \hline
\\[-0.8em]
& FUNIT~\cite{2019FUNIT} \ \                   %
    & 147.3    & 133.6     & 116.9   & 101.7      %
    & 102.7    & 112.4     & 102.9   & 100.3      %
    & 136.8    & 94.9     & 80.9   & 72.6               %
    & 67.6    & 66.5     & \textbf{66.4}   & 67.4               %
\\[0.3mm]
& SEMIT~\cite{wang2020semit} \ \                   %
    & 137.9    & 123.1     & 134.0   & 120.3      %
    & 113.5    & 114.5     & 116.9   & 116.6      %
    & \textbf{74.7}    & 81.7     & \textbf{71.2}   & \textbf{72.5}               %
    & 68.4    & 68.7     & 69.8   & 69.5               %
\\[0.3mm]
& TUNIT (ours)\ \                        
    & \textbf{104.9}    & \textbf{99.9}     & \textbf{96.8}   & \textbf{87.5}      %
    & \textbf{84.0}    & \textbf{79.9}     & \textbf{80.1}   & \textbf{78.4}      %
    & 75.0    & \textbf{74.4}     & 74.9   & 75.2               %
    & \textbf{64.7}    & \textbf{64.0}     & 68.2   & \textbf{66.8}               %
\\[0.3mm] \hline

\end{tabular}\vspace{-2mm}
}
\caption{
\small{Quantitative evaluation (mFID) when few labels are available during training. We note that TUNIT is the same as \texttt{G} in \Tref{tab:main}}
}\vspace{-3mm}
\label{tab:semi_wo_cls}
\end{table*}
}

%% file: 0.abstract.tex
\begin{abstract}
Every recent image-to-image translation model inherently requires either image-level (\ie input-output pairs) or set-level (\ie domain labels) supervision. However, even set-level supervision can be a severe bottleneck for data collection in practice. In this paper, we tackle image-to-image translation in a fully unsupervised setting, \ie, neither paired images nor domain labels. To this end, we propose a truly unsupervised image-to-image translation model (TUNIT) that simultaneously learns to separate image domains and translates input images into the estimated domains. 
Experimental results show that our model achieves comparable or even better performance than the set-level supervised model trained with full labels, generalizes well on various datasets, and is robust against the choice of hyperparameters (\eg the preset number of pseudo domains). Furthermore, TUNIT can be easily extended to semi-supervised learning with a few labeled data. 
\end{abstract}

%% file: 1.introduction.tex
\section{Introduction}
\label{sec:introduction}
 Given an image of one domain, image-to-image translation is a task to generate the plausible images of the other domains. Based on the success of conditional generative models \cite{2014cGAN,2015cVAE}, many image translation methods have been proposed either using \emph{image-level} supervision (\eg paired data)~\cite{2017Pix2Pix,2017Cycada,2017bicyclegan,wang2018pix2pixhd,park2019spade} or using \emph{set-level} supervision (\eg domain labels) \cite{2017CycleGAN,2017DiscoGAN,2017UNIT,2018MUNIT,2019FUNIT,2020DRIT++}. 
Though the latter approach is generally called \emph{unsupervised} as a counterpart of the former, it actually assumes that the domain labels are given \textit{a priori}. This assumption can be a serious bottleneck in practice as the number of domains and samples increases. For example, labeling individual samples of a large dataset (\eg FFHQ) is expensive, and the distinction across domains can be vague.

We first clarify that unsupervised image-to-image translation should strictly denote the task \textit{without any supervision} neither paired images nor domain labels. Under this rigorous definition, our goal is to develop an unsupervised translation model given a mixed set of images of many domains (\Fref{fig:introduction}). We argue that the unsupervised translation model is valuable in three aspects. First of all, it significantly reduces the effort of data annotation for model training. As a natural byproduct, the unsupervised model can be robust against the noisy labels produced by the manual labeling process. More importantly, it serves as a strong baseline to develop the semi-supervised image translation models. To tackle this problem, we design our model having three sub-modules: 1) clustering the images by approximating the set-level characteristics (\ie domains), 2) encoding the individual content and style of an input image, respectively, and 3) learning a mapping function among the estimated domains. 

\figIntro

To this end, we introduce a \texttt{guiding network}. The guiding network consists of a shared encoder with two branches, where one provides pseudo domain labels and the other encodes images into feature vectors (style codes). We employ a differentiable clustering method based on mutual information maximization for estimating the domain labels and contrastive loss for extracting the style codes. The clustering helps the guiding network to group similar images into the same category. Meanwhile, the contrastive loss helps the model to understand the dissimilarity among images and learn better representations. We find that, by solving two tasks together within the same module, both benefit from each other. Specifically, the clustering can exploit rich representations learned by the contrastive loss and improve the accuracy of estimated domain labels. By taking advantage from the clustering module, the style code can also acknowledge the similarity within the same domain, thereby faithfully reflecting the domain-specific nature. 

For both more efficient training and effective learning, we jointly train the guiding network and GAN in an end-to-end manner. 
This allows the guiding network to understand the recipes of domain-separating attributes based on GAN's feedback, and the generator encourages the style code to contain rich information so as to fool the domain-specific discriminator. 
Thanks to these internal and external interactions of the guiding network and GAN, 
our model successfully separates domains and translates images; a truly unsupervised image-to-image translation. 

We quantitatively and qualitatively compare the proposed model with the existing set-level supervised models under unsupervised and semi-supervised settings. The experiments on various datasets show that the proposed model outperforms the baselines over all different levels of supervision. Our ablation study shows that the guiding network helps the image translation model to largely improve the performance. Our contributions are summarized as follows: 
\begin{itemize}[leftmargin=*]
\setlength\itemsep{0mm}
  \item We clarify the definition of unsupervised image-to-image translation and to the best of our knowledge, our model is the first to succeed in this task in an end-to-end manner.
  \item We propose the guiding network to handle the unsupervised translation task and show that the interaction between translation and clustering is helpful for the task.
  \item The quantitative and qualitative comparisons for the unsupervised translation task on four public datasets show the effectiveness of TUNIT, which clearly outperforms the previous arts.
  \item TUNIT is insensitive to the hyperparameter (\ie the number of clusters) and serves as a strong baseline for the semi-supervised setting-- TUNIT outperforms the current state-of-the-art semi-supervised image translation model.
\end{itemize}

%% file: 2.related_work.tex
\section{Related work}
\label{sec:related_work}
\noindent\textbf{Image-to-image translation.} 
Since the seminal work of Pix2Pix \cite{2017Pix2Pix}, image-to-image translation models have shown impressive results~\cite{2017CycleGAN,2017UNIT,2017DiscoGAN,2017Cycada,2018DeblurGAN,2018StarGAN,2018MUNIT,2019FUNIT,2019DSGAN,2019StarGANv2,wang2020semit}. Exploiting the cycle consistency constraint or shared latent space assumption, these methods were able to train the model with a set-level supervision (domains) solely. However, acquiring domain information can be a huge burden in practical applications where a large amount of data are gathered from several mixed domains, \eg, web images \cite{2019Bilion}. Not only does this complicates the data collection, but it restricts the methods only applicable to the existing dataset and domains. S$^3$GAN~\cite{lucic2019high} and Self-conditioned GAN~\cite{liu2020selfconditioned} integrated a clustering method and GAN for high-quality generation using the fewer number or none of the labeled data, respectively. 
Inspired from few shot learning, Liu \textit{et al.}~\cite{2019FUNIT} proposed FUNIT that works on previously unseen target classes. However, FUNIT still requires the labels for training. Wang \textit{et al.}~\cite{wang2020semit} utilized the noise-tolerant pseudo labeling scheme to reduce the label cost at the training process. Recently, Bahng et al.~\cite{bahng2019exploring} partially addressed this by adopting the ImageNet pre-trained classifier for extracting domain information. Unlike the previous methods, we aim to design an image translation model that can be applied without supervision such as a pre-trained network or supervision on both the train and the test datasets. 
\smallskip\\
\noindent\textbf{Unsupervised representation learning and clustering.} Unsupervised representation learning aims to extract meaningful features for downstream tasks without any human supervision. To this end, many researchers have proposed to utilize the information that can be acquired from the data itself~\cite{2006contrastive,2018RotNet,2018DIM,2019IIC,saunshi2019theoretical,2019MoCo,chen2020simclr,2020SCAN}.
Recently, by incorporating contrastive learning into a dictionary learning framework, MoCo~\cite{2019MoCo,2020mocov2} achieved outstanding performance in various downstream tasks under reasonable mini-batch size. 
On the other hand, IIC~\cite{2019IIC} utilized the mutual information maximization in an unsupervised manner so that the network clusters images while assigning the images evenly. 
Though IIC provided a principled way to perform unsupervised clustering, the method fails to scale up when combined with a difficult downstream task such as image-to-image translation. 
By taking the best of both worlds, we aim to solve unsupervised image-to-image translation.

%% file: 3.method.tex
\section{Truly Unsupervised Image-to-Image Translation (TUNIT)}
\label{sec:method}
\figModel

We address the unsupervised image-to-image translation problem, where we have images $\rchi$ from $K$ domains ($K\geq2$) without domain labels $\y$. Here, $K$ is an unknown property of the dataset. Throughout the paper, we denote $K$ as the actual number of domains in a dataset and $\hat{K}$ as the arbitrarily chosen number of domains to train models. 

\subsection{Overview}
In our framework, the guiding network ($E$ in \Fref{fig:model}) plays a central role as an unsupervised domain classifier as well as a style encoder. 
It guides the translation by feeding the style code of a reference image to the generator and its pseudo domain labels to the discriminator. Using the feedback from the domain-specific discriminator, the generator synthesizes an image of the target domain (\eg \ breeds) while respecting the style (\eg \ fur patterns) of the reference image and the content (\eg pose) of the source image.

\subsection{Learning to produce domain labels and encode style features}\label{sec:guiding_encoder}
The guiding network $E$ consists of two branches, $\Eclass$ and $\Estyle$, each of which learns to provide domain labels and style codes, respectively. In experiments, we compare our guiding network against straightforward approaches, \ie., K-means on image or feature space.

\smallskip
\noindent\textbf{Unsupervised domain classification.} 
The discriminator requires a target domain label to provide useful gradients to the generator for translating an image into the target domain. 
Therefore, we adopt the differentiable clustering technique~\cite{2019IIC} that maximizes the mutual information (MI) between an image $\x$ and its randomly augmented version $\xp$. 
The optimum of the mutual information $I(\p, \pp)$ is reached when the entropy $H(\p)$ is maximum and the conditional entropy $H(\p|\pp)$ is minimum, where $\p = \Eclass(\x)$ represents the softmax output 
from $\Eclass$, indicating a probability vector of $\x$ over $\hat{K}$ domains. Please refer to Section~\ref{sec:generalizability} for more details about $\hat{K}$.
Maximizing MI encourages $\Eclass$ to assign the same domain label to the pair ($\x$ and $\xp$) while evenly distributing entire samples to all domains. Formally, $\Eclass$ maximizes the mutual information:
\begin{small}
\begin{equation}
\label{eq:iic}
\begin{split}
\mathcal{L}_{MI} &= I(\p, \pp) = I(\P) = \sum_{i=1}^{\hat{K}} \sum_{j=1}^{\hat{K}} \Pij \ln \frac{\Pij}{\P_i \P_j},\\
\textit{s.t.}\ \ \P&=\mathbb{E}_{\xp\sim f(\x) | \x\sim p_{data}(\x)}[\Eclass(\x)\cdot \Eclass(\xp)^T].
\end{split}
\end{equation}
\end{small}
Here, $f$ is a composition of random augmentations such as random cropping and affine transformation. $\P_i=\P(\p=i)$ denotes the $\hat{K}$-dimensional marginal probability vector, and $\Pij=\P(\p=\texttt{argmax}(i), {\pp}=\texttt{argmax}(j))$ denotes the joint probability. To provide a deterministic one-hot label to the discriminator, we use the \texttt{argmax} operation (i.e. $y = \texttt{argmax}(\Eclass(\x))$). Note that the mutual information is one way to implement TUNIT, and any differentiable clustering methods can be adopted such as SCAN~\cite{2020SCAN}.

\smallskip
\noindent\textbf{Style encoding and improved domain classification.} To perform a reference-guided image translation, the generator needs to understand the style features of the given image. In our framework, $\Estyle$ encodes an image into a style code $\s$, which is later used to guide the generator for image translation. Here, to learn the style representation, we use the contrastive loss~\cite{2019MoCo}:
\begin{small}
\begin{equation}
\label{eq:contrastive}
\mathcal{L}_{style}^E = -\log\frac{\exp(\s\cdot {\spp}/\tau)}{\sum_{i=0}^N \exp(\s\cdot \sn/\tau)},
\end{equation} 
\end{small}
where $\x$ and $\xp$ denote an image and randomly augmented version of $\x$, respectively, and $\s=\Estyle(\x)$.
This $(N+1)$-way classification enables $E$ to utilize not only the similarity of the positive pair ($\s$, $\spp$) but also the dissimilarity of the negative pairs ($\s$, $\sn$).
We adopt a queue to store the negative codes $\sn$ of the previously sampled images as MoCo~\cite{2019MoCo}. By doing so, we can conduct the contrastive learning efficiently without large batch sizes~\cite{saunshi2019theoretical}.

Interestingly, we find that $\Estyle$ also helps the unsupervised domain classification task---$(-\mathcal{L}_{MI} + \mathcal{L}_{style}^E)$ significantly improves the quality of the clustering, compared to using only $-\mathcal{L}_{MI}$, which is the original IIC~\cite{2019IIC}. 
Since $\Estyle$ shares the embeddings with $\Eclass$, imposing the contrastive loss on the style codes improves the representation power of the shared embeddings. 
This is especially helpful when samples are complex and diverse, and IIC solely fails to scale up (\eg, AnimalFaces~\cite{2019FUNIT}). 
To evaluate the effect of Eq.\eqref{eq:contrastive}, we measure the accuracy and the ratio of the inter-variance over the intra-variance (IOI) in terms of the cosine similarity of each clustering result. The clustering result can be more discriminative when the intra-variance and the inter-variance become low and high, respectively. Therefore, the higher IOI indicates a more discriminative clustering result. The table below summarizes the accuracies and IOI on AnimalFaces-10 and Food-10. 
\begin{center}
\vspace{-0.5mm}
\tablestyle{10pt}{1.0}
\begin{tabular}{ccc|cc}
& \multicolumn{2}{c}{\textbf{AnimalFaces-10}} & \multicolumn{2}{c}{\textbf{Food-10}} \\
 & IIC & IIC + Eq.\eqref{eq:contrastive} & IIC & IIC + Eq.\eqref{eq:contrastive} \\
\shline
IOI & 2.05 & 3.04 & 1.34 & 2.50 \\
Accuracy & 68.0\% & 85.0\% & 54.2\% & 86.0\% 
\end{tabular}
\vspace{-0.7mm}
\end{center}
For both datasets, IIC with Eq.\eqref{eq:contrastive} shows a significantly higher accuracy and a higher IOI value. Based on this analysis, we choose to use both $-\mathcal{L}_{MI}$ and $ \mathcal{L}_{style}^E$ for training the guiding network. 
\subsection{Learning to translate images}\label{sec:gan}
We describe how to perform the unsupervised image-to-image translation under the guidance of our guiding network. For successful translation, the model should provide realistic images containing the visual feature of the target domain. To this end, we adopt three losses for training the generator $G$: 1) adversarial loss to produce realistic images, 2) style contrastive loss that encourages the model not to ignore the style codes, 3) image reconstruction loss for preserving the domain-invariant features. We explain each loss and the overall objective for each network.

\smallskip
\noindent\textbf{Adversarial loss.} For adversarial training, we adopt a variant of conditional discriminator, the multi-task discriminator~\cite{mescheder2018r1reg}. It is designed to conduct discrimination for each domain simultaneously. During training, its gradient is calculated only with the loss for estimating the input domain. For the domain label of the input, we utilize the pseudo label from the guiding network. Formally, given the pseudo label $\yt$ for a reference image $\xt$, we train our generator $G$ and multi-task discriminator $D$ via the adversarial loss:
\begin{small}
\begin{equation}
\label{eq:discriminator}
\begin{split}
\mathcal{L}_{adv} &= \mathbb{E}_{\xt\sim p_{data}(\x)}[\log D_{\yt}(\xt)]  \\
& + \mathbb{E}_{\x, \xt \sim p_{data}(\x)}[\log (1 - D_{\yt}(G(\x, \st)))],
\end{split}
\end{equation}
\end{small}
where $D_{\yt}(\cdot)$ denotes the logit from the domain-specific ($\yt$) discriminator, and $\st = \Estyle(\xt)$ denotes a target style code of the reference image $\xt$. The generator $G$ learns to translate $\x$ to the target domain $\yt$ while reflecting the style code $\st$.

\smallskip
\noindent\textbf{Style constrastive loss.} In order to prevent a degenerate case where the generator ignores the given style code $\st$ and synthesize a random image of the domain $\yt$, we impose a style contrastive loss:
\begin{small}
\begin{equation}
\label{eq:stycontrastive}
\mathcal{L}_{style}^G = \mathbb{E}_{\x, \xt \sim p_{data}(\x)}\left[-\log\frac{\exp(\stt \cdot \st)}{\sum_{i=0}^N \exp(\stt \cdot \sn/\tau)}\right].
\end{equation}
\end{small}
Here, $\stt = \Estyle(G(\x, \st))$ denotes the style code of the translated image $G(\x, \st)$ and $\sn$ denotes the negative style codes, which are from the same queue used in Eq.~\eqref{eq:contrastive}. Besides, the same training scheme of MoCo~\cite{2019MoCo} is applied for the generator training as Eq.~\eqref{eq:contrastive}. This loss guides the generated image $G(\x, \st)$ to have a style similar to the reference image $\xt$ and dissimilar to negative (other) samples. By doing so, we avoid the degenerated solution where the encoder maps all the images to the same style code of the reconstruction loss~\cite{2019StarGANv2} based on L1 or L2 norm. Eqs.~\eqref{eq:contrastive} and \eqref{eq:stycontrastive} are based on contrastive loss, but they are used for different purposes. Please refer to Appendix 12. for more discussion.

\smallskip
\noindent\textbf{Image reconstruction loss.} We impose that the generator $G$ can reconstruct the source image $\x$ when given with its original style $\s = \Estyle(\x)$, namely an image reconstruction loss:
\begin{small}
\begin{equation}
\label{eq:reconstruction}
\mathcal{L}_{rec} = \mathbb{E}_{\x\sim p_{data}(\x)}[\lVert \x - G(\x, \s) \rVert_1].
\end{equation}
\end{small}
This objective not only ensures the generator $G$ to preserve domain-invariant characteristics (e.g., pose) of its source image $\x$, but also helps to learn the style representation of the guiding network $E$ by extracting the original style $\s$ of the source image $\x$.

\smallskip
\noindent\textbf{Overall objective.} Finally, we train the three networks jointly as follows:
\begin{small}
\begin{equation}
\label{eq:totobjectives}
\begin{split}
\mathcal{L}_D &= -\mathcal{L}_{adv}, \\
\mathcal{L}_G &= \mathcal{L}_{adv} + \lambda_{style}^G\mathcal{L}_{style}^G + \lambda_{rec}\mathcal{L}_{rec}, \\
\mathcal{L}_E &= \mathcal{L}_G-\lambda_{MI}\mathcal{L}_{MI} + \lambda_{style}^E\mathcal{L}_{style}^E
\end{split}
\end{equation}
\end{small}
where $\lambda$'s are hyperparameters. Note that our guiding network $E$ receives feedback from $L_{G}$, which is essential for our method. We discuss the effect of feedback to $E$ on performance in Section~\ref{sec:comparison}.

%% file: 4.experiments.tex
\section{Experiments}
\label{sec:experiments}
\tabmain
\figmaincomparison
\tabKcomparisonn
\figFFHQfew
\tabsemi
We first investigate the effect of each component of TUNIT~(\Sref{sec:comparison}). We quantitatively and qualitatively compare the performance on labeled datasets. We show that TUNIT is robust against the choice of hyperparameters (e.g. the preset number of clusters, $\tilde{K}$) and extends well to the semi-supervised scenario~(\Sref{sec:generalizability}). Lastly, we move on to unlabeled datasets to validate our model in the unsupervised scenario in the wild~(\Sref{sec:unsupervised}). 


\medskip
\noindent\textbf{Datasets.} 
For the labeled datasets, we select ten classes among 149 classes of AnimalFaces and 101 classes of Food-101, which we call AnimalFaces-10 and Food-10, respectively. Here, the labels are used only for the evaluation purpose except for the semi-supervised setting. For the unlabeled datasets, we use AFHQ, FFHQ, and LSUN Car~\cite{2019StarGANv2,2019StyleGAN,2015LSUN}, which do not have any or are missing with fine-grained labels. Specifically, AFHQ roughly has three groups (\ie, dog, cat and wild), but each group contains diverse species and these species labels are not provided. FFHQ and LSUN Car contain various human faces and cars without any labels, respectively.

\medskip
\noindent\textbf{Evaluation metrics.} We report two scores to assess the generated images. First, to provide a general sense of image quality, we use the mean of class-wise Fr\'enchet Inception Distance (mFID)~\cite{2017FID}. It can avoid the degenerate case of the original FID, which assigns a good score when the model conveying the source image as is. Additionally, to provide a finer assessment of the generated images, we report Density and Coverage (D\&C)~\cite{2020DnC}. D\&C separately evaluates the fidelity and the diversity of the model outputs, which is also known to be robust against outliers and model hyperparameters (e.g. the number of samples used for evaluation). A lower mFID score means better image quality, and D\&C scores that are bigger or closer to 1.0 indicate better fidelity and diversity, respectively. Please refer to Appendix 3. for the detailed information.
\subsection{Comparative Evaluation on Labeled Datasets}
\label{sec:comparison}
\Tref{tab:main} summarizes the effect of each component of TUNIT and rigorous comparisons with the state-of-the-art supervised method, FUNIT. First, we report the set-level supervised performance of FUNIT and its variant (\Tref{tab:main}). Here, \texttt{A} is the original FUNIT and \texttt{B} denotes the modified FUNIT using our architecture (\eg we do not use PatchGAN discriminator), which brings a large improvement over every score on both datasets. 
One simple way to extend \texttt{B} to the unsupervised scenario is to add an off-the-shelf clustering method and use its estimated labels instead of the ground truth. We employ K-means clustering on the image space for \texttt{C}, and the pretrained feature space for \texttt{D}. Here, we use ResNet-50~\cite{2016resnet} features trained with MoCo v2~\cite{2020mocov2} on ImageNet. Not surprisingly, because the estimated labels are inaccurate, the overall performance significantly drops. 
Although using the pretrained features helps a little, not only is it far from the set-level supervised performance but it requires three steps to train the entire model, which complicates the application. This can be partially addressed by employing the differentiable clustering method~\cite{2019IIC}, which trains VGG-11BN~\cite{2014VGG} with mutual information maximization from scratch that makes \texttt{E}. This reduces the number of training steps from three to two and provides better label estimation, which enables the model to approach the performance of original FUNIT \texttt{A}. However, as seen in the coverage score, the sample diversity is unsatisfactory. 

Finally, we build TUNIT by introducing the guiding network and the new objective functions described in \Sref{sec:method}. The changes significantly improve the accuracy on both datasets, particularly achieving similar mFID of the improved set-level supervised model \texttt{B}. Our final model, \texttt{G} matches or outperforms mFID and D\&C of \texttt{B}. This is impressive because \texttt{B} utilizes oracles for training while \texttt{G} has no labels. Notably, TUNIT can improve the coverage by 0.073 (7\%p) on AnimalFaces-10 than \texttt{B}. 
We conjecture that TUNIT benefits from the guiding network, which jointly learns the style encoding and clustering with the shared encoder. Because their clustering modules are as powerful as TUNIT, the performance drawback of \texttt{C}, \texttt{D} and \texttt{E} supports that the feedback from style encoding is a key success factor of TUNIT. By comparing \texttt{F} and \texttt{G}, we confirm that they are comparable in terms of clustering and \texttt{G} is more stable in terms of inter-dataset performance. Therefore, we adopt the joint training of style encoder and clustering as our final model (\texttt{G}). In addition, we investigate the effect of joint training between GAN and the guiding network by removing the adversarial loss for training the guiding network. It directly degrades the performance; mFID changes from 47.7 to 63.0 on AnimalFaces-10. It indicates that our training scheme takes an important portion of performance gains. Qualitative results also show the superiority of TUNIT over competitors~(\Fref{fig:main_comparison}).

\figUnlabeled
\figtsnecompare
\subsection{Analysis on Generalizability}
\label{sec:generalizability}
\noindent\textbf{Robustness to various $\hat{K}$'s.} When TUNIT conducts clustering for estimating domain labels, the number of clusters $\hat{K}$ can affect the performances. Here, we study the effects on different $\hat{K}$ on the labeled datasets and report them in \Tref{tab:k_comparison}. For the qualitative comparison, please refer to Appendix 1. As expected, the model performs best in terms of mFID when $\hat{K}$ equals to the ground truth $K$ (\ie $\hat{K}$=10). 

One thing to note here is that TUNIT performs reasonably well for a sufficiently large $\hat{K}$ ($\geq 7$). More interestingly, even with 100 times larger $\hat{K}$ than the actual number of the domains, TUNIT still works well on both datasets. This trend is also seen in the t-SNE visualization(~\Tref{tab:k_comparison}). From this study, we conclude that TUNIT is relatively robust against $\hat{K}$ as long as it is sufficiently large. Thus, in practice, we suggest to use a sufficiently large $\hat{K}$ or to study different $\hat{K}$'s in log scale for finding the optimal model. 

\medskip
\noindent\textbf{With Few labels.} We also investigate whether or not TUNIT is effective for a more practical scenario, semi-supervised image translation. To utilize the labels, AnimalFaces and Food datasets are chosen for this experiment. Specifically, AnimalFaces-10 and Food-10 are used for evaluating the models on the small datasets while AnimalFaces-149 and Food-101 are used for assessing the models on the large datasets. We partition the dataset $\mathcal{D}$ into the labeled set $\mathcal{D}_{sup}$ and the unlabeled set $\mathcal{D}_{un}$ with varying ratio $\gamma=|\mathcal{D}_{sup}|/|\mathcal{D}|$. For the semi-supervising setting, we train $\Eclass$ of TUNIT by an additional cross-entropy loss between the ground truth domain labels and the predicted domain labels on $\mathcal{D}_{sup}$. Besides, the true domain labels for $\mathcal{D}_{sup}$ are utilized for training the domain-specific discriminator. 
As a counterpart in this scenario, FUNIT~\cite{2019FUNIT} and SEMIT~\cite{wang2020semit} are selected because both models can be applied to the semi-supervised image translation (SEMIT achieves the current state-of-the-art performance in the semi-supervised setting). We train the two competitors and TUNIT by changing $\gamma$ from 0.01 to 0.8 and report the results in \Tref{tab:semi_wo_cls}. For the small datasets, the performance of FUNIT significantly degrades as $\gamma$ decreases. Meanwhile, TUNIT and SEMIT produce relatively similar mFID scores despite $\gamma$ decreases. Even when SEMIT maintains mFID, TUNIT significantly outperforms SEMIT by 20\% of mFID on small datasets. For the large datasets, TUNIT either outperforms or is comparable to the competitors. Especially, on AnimalFaces-149, TUNIT clearly outperforms both competitors. By the experiments on the semi-supervised setting, we conclude that TUNIT can be easily adapted to the semi-supervised image translation with the simple modification (\ie adding the supervised training on the labeled samples), and serve as a strong baseline model. As a result, TUNIT for the semi-supervised setting achieves impressive performance, which is comparable to the state-of-the-art semi-supervised translation method. We provide the qualitative comparison in the appendix.

\subsection{Validation on Unlabeled Dataset}
\label{sec:unsupervised}

Finally, we evaluate TUNIT on the unlabeled datasets (AFHQ, FFHQ and LSUN-Car), having no clear separations of the domains. For AFHQ, we train three individual models for \emph{dog}, \emph{cat} and \emph{wild}. For all experiments, we use FUNIT as a baseline, where all the labels for training are regarded the same as one. We set the number of clusters $\hat{K}$=10 for all the TUNIT models. 

Figure~\ref{fig:unlabeled_reference_guided} demonstrates the results. We observe that the results of TUNIT adequately reflect the style feature of the references such as the textures of cats or cars and the species of the wilds. Although FFHQ has no clear domain distinctions, TUNIT captures the existence of glasses or smile as domains, and then add or remove glasses or smile. However, FUNIT performs much worse than TUNIT in this truly unsupervised scenario. For example, FUNIT outputs the inputs as is (cats and wilds) or insufficiently reflects the species (third row of AFHQ Wild). For FFHQ, despite that FUNIT makes some changes, the changes are not interpreted as meaningful domain translations. For LSUN Car, FUNIT fails to keep the fidelity. 

We also visualize the style space of both models to qualitatively assess the quality of the representation. \Fref{fig:tsne_compare} shows the t-SNE maps trained on AFHQ Wild and the examples of each cluster; the sample color corresponds to the box color of representative images. Surprisingly, TUNIT organizes the samples according to the species where it roughly separates the images into six species. Although we set $\hat{K}$ to be overly large, the model represents one species into two domains where those two domains position much closely (\eg tiger). From these results, we confirm that the highly disentangled, meaningful style features can be an important factor in the success of our model. 
On the other hand, the style features of FUNIT hardly learn meaningful domains so that the model cannot conduct the translation properly as shown in \Fref{fig:unlabeled_reference_guided}. 
Because of the page limit, we include more results including qualitative comparison and the t-SNE visualization in Appendix 8,9,10 and 11.

%% file: 5.conclusion.tex
\section{Conclusion}
\label{sec:conclusion}
We argue that an unsupervised image translation should denote a task that does not utilize any kinds of supervision, neither image-level (\ie paired) nor set-level (\ie unpaired). In this regime, most of the previous studies fall into the set-level supervised framework, using the domain information at a minimum. 
In this paper, we proposed TUNIT, a truly unsupervised image translation method. 
By exploiting synergies between clustering and representation learning, TUNIT finds pseudo labels and style codes so that it can translate images without using any external information. 
The experimental results show that TUNIT can successfully perform an unsupervised image translation while being robust against hyperparameter changes (\eg, the preset number of clusters, $\hat{K}$). Our model is easily extended to the semi-supervised setting, providing comparable results to the state-of-the-art semi-supervised method. These highlight TUNIT has great potential in practical applications.


%% file: 6.appendix.tex
\appendix
\newcommand{\supplesection}{\section}
\newcommand{\supplesubsection}{\subsection}
\figKcomparison
\supplesection{Qualitative comparison on the number of pseudo domains $\hat{K}$}
\label{sec:k_comparison}
Please refer to Figure 1. in Appendix.
\supplesection{Training details}
\label{sec:app_training}
We train the guiding network for the first 65K iterations while freezing the update from both the generator and the discriminator. Then, we train the whole framework 100K more iterations for training all the networks. The batch size is set to 32 and 16 for 128$\times$128 and 256$\times$256 images, respectively. Training takes about 36 hours on a single Tesla V100 GPU with our implementation using PyTorch\cite{paszke2017pytorch}.
We use Adam~\cite{kingma2014adam} optimizer with $\beta_1=0.9, \beta_2=0.99$ for the guiding network, and RMSprop~\cite{hinton2012rmsprop} optimizer with $\alpha=0.99$ for the generator and the discriminator. All learning rates are set to $0.0001$ with a weight decay $0.0001$.
We adopt hinge version adversarial loss~\cite{lim2017hinge,tran2017hinge} with $R_1$ regularization~\cite{mescheder2018r1reg} using $\gamma=10$ (Eq. 5). 
We set $\lambda_{\text{rec}}=0.1, \lambda_{\text{style}}^G=0.01, \lambda_{\text{style}}^E=1,$ and  $\lambda_{\text{MI}}=5$ in equation. 6 for all experiments. 
When the guiding network is simultaneously trained with the generator, we decrease $\lambda_{\text{style}}^E$ and $\lambda_{\text{MI}}$ to 0.1 and 0.5, respectively.
For evaluation, we use the exponential moving average over the parameters~\cite{karras2017progressive} of the guiding network and the generator.
We initialize the weights of convolution layers with He initialization~\cite{he2015delving}, all biases to zero, and weights of linear layers from $N(0, 0.01)$ with zero biases. The source code will be available publicly. The source code is available at \url{https://github.com/clovaai/tunit}.
\supplesection{Evaluation protocol}
\label{sec:evaluation_protocol}
For evaluation, we use class-wise Fr\'echet Inception Distance (FID)~\cite{2017FID}, which is often called mFID in literatures and D\&C~\cite{2020DnC}. FID measures Fr\'echet distance between real and fake samples embedded by the last average pooling layer of Inception-V3 pre-trained on ImageNet. Class-wise FID is obtained by averaging the FIDs of individual classes. In the experiments with fewer labels, we report the mean value of best five mFID's over 100K iterations.
For example, we use entire real images of each class and generate 810 fake images where $18 \times (K - 1)$ source images ($K=10$ for AnimalFaces-10) and five reference images of AnimalFaces-10 are used to produce those fake images. We choose the source images from all classes except for the target class. For each source image, the five references are selected arbitrarily. For D\&C, we generate fake images the similar number of training images with randomly selected source and reference images. Then, we use Inception-V3 pre-trained on ImageNet for extracting feature vectors and measure D\&C by using the feature vectors.

\newcommand{\shape}[3]{${#1}\times{#2}\times{#3}$}

\supplesection{Difference between the sequential and joint training in Section 4.1}
To investigate the effect of the adversarial loss to the guiding network, we trained TUNIT under two settings; 1) joint training and 2) sequential training. The former is to train all the networks in an end-to-end manner as described in Section 3, and the latter is to first train the guiding network with
$\mathcal{L}_E$ for 100k iterations and then train the generator and the discriminator using the outputs of the frozen guiding network as their inputs. Note that for the separate training, the guiding network does not receive feedback from
the translation loss $\mathcal{L}_G$ in Eq. (6).

\supplesection{Architecture details}
For the guiding network, we use \texttt{VGG11} before the linear layers followed by the average pooling operation as the shared part and append two branches $E_\text{class}$ and $E_\text{style}$. The branches are one linear layer with $\hat{K}$ and $128$ dimensional outputs, respectively. The detailed information of the generator, the guiding network and the discriminator architectures are provided in \Tref{tab:G_arch}, \Tref{tab:Guiding_arch} and \Tref{tab:D_arch}.

\begin{table}[H]
\centering
\begin{tabular}{lccc}
\toprule
\textsc{Layer} & \textsc{Resample} & \textsc{Norm} & \textsc{Output Shape}\vspace{0.5mm}\\
\toprule 
Image $\x$ & - & - & \shape{128}{128}{3}\\
\midrule
Conv7$\times$7 & - & IN & \shape{128}{128}{ch}\\
Conv4$\times$4 & Stride 2 & IN & \shape{64}{64}{2ch}\\
Conv4$\times$4 & Stride 2 & IN & \shape{32}{32}{4ch}\\
Conv4$\times$4 & Stride 2 & IN & \shape{16}{16}{8ch}\\
\midrule
ResBlk & - & IN & \shape{16}{16}{8ch}\\
ResBlk & - & IN & \shape{16}{16}{8ch}\\
ResBlk & - & AdaIN & \shape{16}{16}{8ch}\\
ResBlk & - & AdaIN & \shape{16}{16}{8ch}\\
\midrule
Conv5$\times$5 & Upsample & AdaIN & \shape{32}{32}{4ch}\\
Conv5$\times$5 & Upsample & AdaIN & \shape{64}{64}{2ch}\\
Conv5$\times$5 & Upsample & AdaIN & \shape{128}{128}{ch}\\
Conv7$\times$7 & - & - & \shape{128}{128}{3}\\
\bottomrule
\end{tabular}\vspace{2mm}
\caption{Generator architecture. ``ch'' represents the channel multiplier that is set to 64. IN and AdaIN indicate instance normalization and adaptive instance normalization, respectively.}
\label{tab:G_arch}
\end{table}

\begin{table}[H]
\centering
\begin{tabular}{lccc}
\toprule
\textsc{Layer} & \textsc{Resample} & \textsc{Norm} & \textsc{Output Shape}\vspace{0.5mm}\\
\toprule 
Image $\x$ & - & - & \shape{128}{128}{3}\\
\midrule
Conv3$\times$3 & MaxPool & BN & \shape{64}{64}{ch}\\
Conv3$\times$3 & MaxPool & BN & \shape{32}{32}{2ch}\\
Conv3$\times$3 & - & BN & \shape{32}{32}{4ch}\\
Conv3$\times$3 & MaxPool & BN & \shape{16}{16}{4ch}\\
Conv3$\times$3 & - & BN & \shape{16}{16}{8ch}\\
Conv3$\times$3 & MaxPool & BN & \shape{8}{8}{8ch}\\
Conv3$\times$3 & - & BN & \shape{8}{8}{8ch}\\
Conv3$\times$3 & MaxPool & BN & \shape{4}{4}{8ch}\\
\midrule
GAP & - & - & \shape{1}{1}{8ch}\\
FC & - & - & 128\\
FC & - & - & $\hat{K}$\\
\bottomrule
\end{tabular}\vspace{2mm}
\caption{Guiding network architecture. ``ch'' represents the channel multiplier that is set to 64. The architecture is based on VGG11-BN. GAP and FC denote global average polling~\cite{lin2013network} and fully connected layer, respectively.}
\label{tab:Guiding_arch}
\end{table}

\begin{table}[H]
\centering
\begin{tabular}{lccc}
\toprule
\textsc{Layer} & \textsc{Resample} & \textsc{Norm} & \textsc{Output shape}\vspace{0.5mm}\\
\toprule 
Image $\x$ & - & - & \shape{128}{128}{3}\\
\midrule
Conv3$\times$3 & - & - & \shape{128}{128}{ch}\\
ResBlk & - & FRN & \shape{128}{128}{ch}\\
ResBlk & AvgPool & FRN & \shape{64}{64}{2ch}\\
ResBlk & - & FRN & \shape{64}{64}{2ch}\\
ResBlk & AvgPool & FRN & \shape{32}{32}{4ch}\\
ResBlk & - & FRN & \shape{32}{32}{4ch}\\
ResBlk & AvgPool & FRN & \shape{16}{16}{8ch}\\
ResBlk & - & FRN & \shape{16}{16}{8ch}\\
ResBlk & AvgPool & FRN & \shape{8}{8}{16ch}\\
ResBlk & - & FRN & \shape{8}{8}{16ch}\\
ResBlk & AvgPool & FRN & \shape{4}{4}{16ch}\\
\midrule
LReLU & - & - & \shape{4}{4}{16ch}\\
Conv4$\times$4 & - & - & \shape{1}{1}{16ch}\\
LReLU & - & - & \shape{1}{1}{16ch}\\
\midrule
Conv1$\times$1 & - & - & $\hat{K}$\\
\bottomrule
\end{tabular}\vspace{2mm}
\caption{Discriminator architecture. ``ch'' and $\hat{K}$ represent the channel multiplier that is set to 64 and the number of clusters, respectively. FRN indicates filter response normalization~\cite{singh2019frn}.}
\label{tab:D_arch}
\end{table}

\supplesection{Comparison with Swapping autoencoder}
\begin{figure}[H]
\includegraphics[width=0.99\columnwidth]{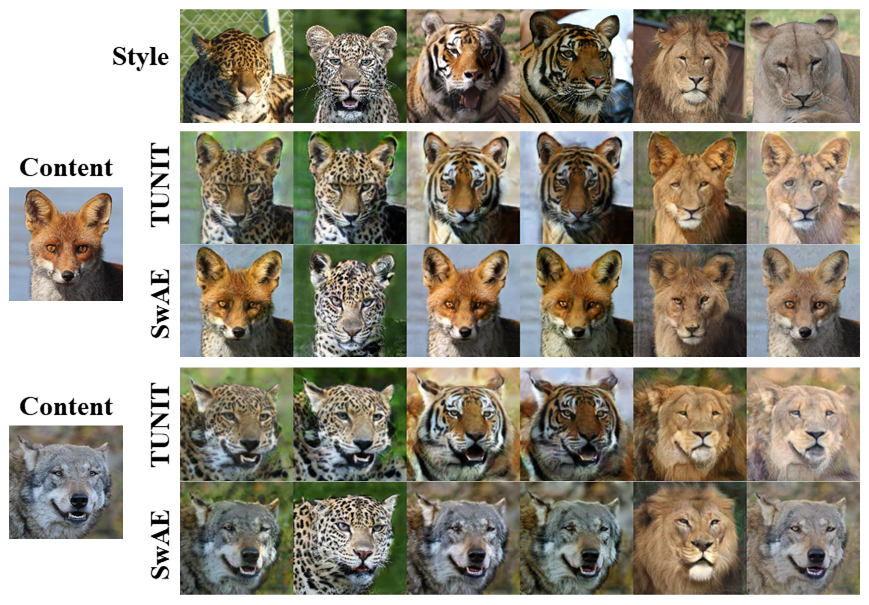}
    \caption{\small{Comparison with SwAE. SwAE sometimes fails to capture the domain features. We recommend to zoom in.}} \label{fig:vsSwAE}
\vspace{-3mm}
\label{fig:vsSwap}
\end{figure}
Swapping autoencoder (SwAE)~\cite{park2020swapping} can conduct the image translation without the domain labels by using the feature vector of the reference images. Therefore, it can be compared with TUNIT. \Fref{fig:vsSwap} shows the qualitative comparison between TUNIT and SwAE. Since SwAE does not define domains, it occasionally fails to capture the exact domain properties (col. 1,3,4 and 6). Meanwhile, TUNIT captures various aspects of domains; it changes species along with styles. It shows that TUNIT better handles translation across domains. Originally, we did not compare SwAE because their practical usefulness and the possible tasks differ from ours. Notably, similar to many unsupervised learners, TUNIT can serve as a strong baseline for semi-supervised models. Table 3 and Fig. 4 show that TUNIT is successful
in a semi-supervised setting. This is clearly not possible by SwAE due to their design choice. Besides, TUNIT further translates domains specified by cluster ids via their average
styles (Appendix Fig. 10). It is especially useful when a user wants to explore various domains without references. In contrast, SwAE always requires a reference.

\supplesection{Comparison with StarGANv2}
We additionally compare TUNIT with StarGANv2 on AnimalFaces-10 and Food-10. We employ StarGANv2 with supervision as the reference for the upper-bound. The table below shows the quantitative result.
\begin{center}
\footnotesize
\vspace{-1mm}
\tablestyle{10pt}{1.0}
\begin{tabular}{m{28mm}m{2mm}m{8mm}|m{2mm}m{8mm}}
& \multicolumn{2}{c}{\textbf{AnimalFaces-10}} & \multicolumn{2}{c}{\textbf{Food-10}} \\
 & mFID & D\&C & mFID & D\&C \\
\shline
StarGANv2 \scriptsize{(Supervised)} & 33.67 & 1.54/0.91 & 65.03 & 1.09/0.76 \\
TUNIT \scriptsize{(Unsupervised)} & 47.70 & 1.04/0.81 & 52.20 & 1.08/0.87
\end{tabular}
\vspace{-1mm}
\end{center}
StarGANv2 outperformed unsupervised TUNIT on AnimalFaces-10, but TUNIT outperformed StarGANv2 on Food-10. Considering that TUNIT uses no labels and StarGANv2 uses set-level labels, we emphasize that our achievement is impressive.

\supplesection{Perceptual study on disentanglement}
We conducted the user study (selecting the best in style and content) on two datasets and compared models (from Table 1) as follows. The result shows that the proposed (\texttt{F,G}) largely outperforms the others. 
\begin{center}
\footnotesize
\vspace{-2mm}
\tablestyle{10pt}{1.0}
\begin{tabular}{m{14mm}m{2mm}m{2mm}m{2mm}m{2mm}m{2mm}|m{2mm}m{2mm}}
& \texttt{A} & \texttt{B} & \texttt{C} & \texttt{D} & \texttt{E} & \texttt{F} & \texttt{G} \\
\shline
Preference(\%)$\uparrow$ & 3.0 & 6.2 & 3.2 & 10.8 & 12.3 & 19.2 & 45.3\\
\end{tabular}
\vspace{-2mm}
\end{center}

\clearpage

\supplesection{t-SNE visualization \& cluster example images}
\label{sec:app_tsne}
\vspace{5mm}
\figTsneAfhqCat
\newpage
\vspace{5mm}
\figTsneAfhqDog
\clearpage
\vspace{5mm}
\vspace{5mm}
\figTsneFfhq
\newpage
\vspace{5mm}
\figTsneLsun
\clearpage

\supplesection{Additional Comparison with FUNIT: AFHQ, LSUN Car and FFHQ}
\label{sec:app_additional_vs_funit}
\figAppCatVsFUNIT
\figAppWildVsFUNIT
\newpage
\figAppLSUNVsFUNIT
\figAppFFHQVsFUNIT
\clearpage

\supplesection{Additional Results of TUNIT including semi-supervised setting}
\label{sec:app_additional_examples}
\subsection{AnimalFaces-10}
\figAppAnimalFaces
\clearpage

\subsection{AFHQ Cat}
\figAppAFHQCat
\clearpage

\subsection{AFHQ Dog}
\figAppAFHQDog
\clearpage

\subsection{AFHQ Wild}
\vspace{5mm}
\figAppAFHQWild
\clearpage

\subsection{FFHQ} 
\vspace{5mm}
\figAppFFHQ
\clearpage

\subsection{LSUN Car} 
\vspace{5mm}
\figAppLSUN
\clearpage

\subsection{Summer2Winter (S2W)} 
\vspace{5mm}
\figAppStoW
\clearpage

\subsection{Photo2Ukiyoe} 
\vspace{5mm}
\figAppPtoU
\clearpage

\subsection{AnimalFaces-149, comparing with SEMIT} 
\vspace{5mm}
\figAppSEMIT
\clearpage

\supplesection{Difference between equation (2) and equation (4)}
\label{sec:app_eq2_eq4}
Equation (2) and (4) have similar forms -- contrastive loss, but they are used for different purposes. We use equation (2) to improve the representation power of the guiding network, which affects the performance of the generator and the discriminator. On the other hand, equation (4) is used to enforce the generator to reflect the style of a reference image when translating a source image. To examine the effect of each loss, we train models without either equation (2) or (4) on AnimalFaces-10. The mFID score without equation (2) or (4) is 86.8 and 93.3, respectively. Both models are significantly worse than the original setting (47.7). It means that both equation (2) and (4) should be considered during training. 
In addition to the purpose, they are different in terms of the way to choose positive pairs. We use a real image and its randomly augmented version as a positive pair in equation (2) while we use the translated image and reference image as a positive pair. 
In summary, the role of equation (2) is to enhance the representation power of the guiding network and lead the guiding network to learn how to encode the style vector in terms of a style encoder while the role of equation (4) is to guide the generator to learn how to interpret the provided style vector as a form of the output image.

\supplesection{FID and LPIPS on unlabeled dataset}
\label{sec:app_lpips}
\figAppLPIPS
We also utilize LPIPS to evaluate the models in addition to FID and D\&C. However, LPIPS is not proper to evaluate the loyalty for reflecting the reference image and the fidelity of images, we use LPIPS with FID. \Fref{fig:AppLPIPS} shows the result. It is clear that a model with high FID and LPIPS generates a noise-like image. Even if FID is low, a model with high LPIPS also fails to conduct the reference-guided image translation, because it does not preserve the structure of the source image. The model with low LPIPS and high FID might be an adversarial example of LPIPS. We generate the image via optimization on LPIPS. If a model exhibits low FID and LPIPS, it might not reflect the visual feature of the reference image enough. The simple combination of LPIPS and FID can detect several failed models but can not evaluate the loyalty for the reference image. We suggest that the rigorous way to combine several metrics for the quantitative evaluation of the reference-guided translation might be a interesting future work.